\documentclass[twoside]{article}

\usepackage[accepted]{aistats2021}

\usepackage[round]{natbib}

\usepackage[utf8]{inputenc} %
\usepackage[T1]{fontenc}    %
\usepackage[colorlinks=true,citecolor=black,urlcolor=black,linkcolor=black]{hyperref}
\usepackage{url}            %
\usepackage{booktabs}       %
\usepackage{amsfonts}       %
\usepackage{nicefrac}       %
\usepackage{microtype}      %
\usepackage{subcaption}
\usepackage{verbatim}

\usepackage{amsmath,amssymb}
\usepackage[capitalize,nameinlink]{cleveref}
\usepackage{xspace}

\crefname{section}{Sec.}{Secs.}
\crefname{proposition}{Prop.}{Props.}
\crefname{lemma}{Lem.}{Lems.}
\crefname{model}{Mod.}{Mods.}
\crefname{appendix}{App.}{Apps.}

\usepackage{bm}

\usepackage{xspace}
\newcommand{\eg}{\textit{e.g.}\xspace}
\newcommand{\ie}{\textit{i.e.}\xspace}

\renewcommand{\boldsymbol}[1]{\mathbf{#1}} %
\newcommand{\dd}{\,\mathrm{d}} %
\newcommand{\vB}{ \boldsymbol{B} }
\newcommand{\vC}{ \boldsymbol{C} }

\newcommand{\vF}{ \boldsymbol{F} }
\newcommand{\vT}{ \boldsymbol{T} }

\newcommand{\vH}{ \boldsymbol{H} }

\newcommand{\vR}{ \boldsymbol{R} }
\newcommand{\vA}{ \boldsymbol{A} }
\newcommand{\vP}{ \boldsymbol{P} }
\newcommand{\vQ}{ \boldsymbol{Q} }
\newcommand{\vL}{ \boldsymbol{L} }
\newcommand{\vM}{ \boldsymbol{M} }

\newcommand{\vX}{ \boldsymbol{X} }
\newcommand{\vZ}{ \boldsymbol{z} }

\newcommand{\vW}{ \boldsymbol{W} }
\newcommand{\vs}{ \boldsymbol{s} }
\newcommand{\vcs}{ \bm{s} }

\newcommand{\vg}{ \boldsymbol{g} }

\newcommand{\vr}{ \boldsymbol{r} }
\newcommand{\vf}{ \boldsymbol{f} }
\newcommand{\vx}{ \boldsymbol{x} }

\newcommand{\vm}{ \boldsymbol{m} }

\newcommand{\vK}{ \boldsymbol{K} }
\newcommand{\vu}{ \boldsymbol{u} }
\newcommand{\vw}{ \boldsymbol{w} }
\newcommand{\vy}{ \boldsymbol{y} }
\newcommand{\vz}{ \boldsymbol{z} }
\newcommand{\vv}{ \boldsymbol{v} }
\newcommand{\veps}{ \bm{\varepsilon} }
\newcommand{\vq}{ \boldsymbol{q} }
\newcommand{\vPhi}{ \boldsymbol{\Phi} }
\newcommand{\vphi}{ \bm{\phi} }
\newcommand{\vmu}{ \bm{\mu} }

\newcommand{\vSigma}{ \boldsymbol{\Sigma} }

\newcommand{\vlambda}{ \bm {\lambda} }
\newcommand{\vLambda}{ \boldsymbol {\Lambda} }
\newcommand{\vDelta}{ \boldsymbol {\Delta} }

\newcommand{\bigO}{ {\cal O}}

\newcommand{\NN}{ {\cal N} }
\newcommand{\tNN}{ \tilde{{\cal N}} }

\newcommand{\XX}{ {\cal X} }
\newcommand{\cZ}{ {\cal Z} }
\newcommand{\cQ}{{\cal Q}}
\newcommand{\cP}{{\cal P}}

\newcommand{\EE}{\mathbb{E}}
\newcommand{\cL}{{\cal L}}

\newcommand{\GP}{ \mathcal{GP} }
\renewcommand{\mid}{\,|\,} %
\newcommand{\bl}{\backslash}

\newcommand{\om}{{m}}
\newcommand{\op}{{m+1}}

\usepackage{mathtools}
\DeclarePairedDelimiterX{\infdivx}[2]{(}{)}{%
  #1\;\delimsize\|\;#2%
}

\newcommand{\diff}[2]{\mathrm{\frac{\partial\mathit{#1}}{\partial\mathit{#2}}}}
\newcommand{\diffII}[2]{\mathrm{\frac{\partial^2\mathit{#1}}{\partial\mathit{#2}^2}}}

\newcommand{\ssvgp}{S$^2$VGP\xspace}
\newcommand{\sspep}{S$^2$PEP\xspace}
\newcommand{\sscvi}{S$^2$CVI\xspace}
\newcommand{\sspl}{S$^2$PL\xspace}
\newcommand{\sseks}{S$^2$EKS\xspace}

\renewcommand{\ss}{\boldsymbol{\mathsf{s}}}

\usepackage{tikz,pgfplots}
\usetikzlibrary{plotmarks,shapes,shapes.arrows,positioning,calc,spy,intersections}
\pgfplotsset{compat=newest} 
\pgfkeys{/pgf/number format/.cd,1000 sep={}}
\usepackage{placeins}

\renewcommand{\vec}[1]{\mathbf{#1}}
\definecolor{blue}{rgb}{0.12156862745098039, 0.4666666666666667, 0.7058823529411765}
\definecolor{red}{rgb}{0.8392156862745098, 0.15294117647058825, 0.1568627450980392}

\newlength\figureheight
\newlength\figurewidth

\newcommand{\RR}{ \mathbb{R} }
\newcommand{\mprod}{\textstyle\prod}
\newcommand{\mint}{\textstyle\int}
\newcommand{\msum}{\textstyle\sum}

\usepackage{tabularx}
\usepackage{array}
\newcommand{\PreserveBackslash}[1]{\let\temp=\\#1\let\\=\temp}
\newcolumntype{C}[1]{>{\PreserveBackslash\centering}p{#1}}

\usepackage{tikz,pgfplots}
\usetikzlibrary{plotmarks,shapes,arrows}
\pgfplotsset{compat=newest} 

\pgfplotsset{/pgf/number format/.cd, 1000 sep={}}
\pgfkeys{/pgf/number format/.cd,1000 sep={}}

\pgfplotsset{every axis/.append style={
		grid style={line width=0.6pt,dotted,gray}}}

\pgfplotsset{every axis/.append style={
		legend style={inner xsep=1pt, inner ysep=0.5pt, nodes={inner sep=1pt, text depth=0.1em},draw=none,fill=none}
}}

\pgfplotsset{ignore legend/.style={every axis legend/.code={\let\addlegendentry\relax}}}

\usepgfplotslibrary{groupplots}

\renewcommand{\paragraph}[1]{{\bf#1}~~}

\author{%
  David S.~Hippocampus\thanks{Use footnote for providing further information
    about author (webpage, alternative address)---\emph{not} for acknowledging
    funding agencies.} \\
  Department of Computer Science\\
  Cranberry-Lemon University\\
  Pittsburgh, PA 15213 \\
  \texttt{hippo@cs.cranberry-lemon.edu} \\
}

\begin{document}

\twocolumn[

\aistatstitle{Sparse Algorithms for Markovian Gaussian Processes}

\aistatsauthor{ 
William J. Wilkinson  
\And 
Arno Solin 
\And 
Vincent Adam  
}

\aistatsaddress{ \texttt{william.wilkinson@aalto.fi} \\ Aalto University \\ The Alan Turing Institute \And  \texttt{arno.solin@aalto.fi} \\ Aalto University \And \texttt{vincent.adam@secondmind.ai} \\ Secondmind.ai } ]

\begin{abstract}

Approximate Bayesian inference methods that scale to very large datasets are crucial in leveraging probabilistic models for real-world time series. Sparse Markovian Gaussian processes combine the use of inducing variables with efficient Kalman filter-like recursions, resulting in algorithms whose computational and memory requirements scale linearly in the number of inducing points, whilst also enabling parallel parameter updates and stochastic optimisation. Under this paradigm, we derive a general \emph{site-based} approach to approximate inference, whereby we approximate the non-Gaussian likelihood with local Gaussian terms, called sites. Our approach results in a suite of novel sparse extensions to algorithms from both the machine learning and signal processing literature, including variational inference, expectation propagation, and the classical nonlinear Kalman smoothers. The derived methods are suited to large time series, and we also demonstrate their applicability to spatio-temporal data, where the model has separate inducing points in both time and space.

\end{abstract}

\section{INTRODUCTION}
\label{sec:introduction}
Gaussian processes \citep[GPs,][]{rasmussen2003gaussian} are distributions over functions, commonly used in probabilistic machine learning to endow latent functions in generative models with rich and interpretable priors. These priors provide strong inductive biases for regression tasks in the small data regime.

\begin{figure}[hbt!]
  \scriptsize
  \centering
  \pgfplotsset{yticklabel style={rotate=90}, ylabel style={yshift=0pt}, xlabel style={yshift=4pt},scale only axis,axis on top,clip=true,clip marker paths=true}
  \pgfplotsset{legend style={inner xsep=1pt, inner ysep=1pt, row sep=0pt},legend style={at={(0.98,0.95)},anchor=north east},legend style={rounded corners=1pt}}
  \setlength{\figurewidth}{.475\textwidth}
  \setlength{\figureheight}{.44\figurewidth}
  \begin{subfigure}[b]{\columnwidth}
    \centering
    \input{./graphs/teaser_filter.tex}
    \vspace{-0.5cm}
    \caption{$\longrightarrow$ Forward {\em filtering} pass}
    \label{fig:filter}  
  \end{subfigure}
  \\[0.6em]
  \begin{subfigure}[b]{\columnwidth}
    \centering
    \input{./graphs/teaser_smoother.tex}
    \vspace{-0.45cm}
    \caption{$\longleftarrow$ Backward {\em smoothing} pass}
    \label{fig:smoother}  
  \end{subfigure}
  \vspace*{-0.5cm}
  \caption{Inducing states with Markovian GPs: filtering \textbf{(a)} and smoothing (\ie, posterior) distribution \textbf{(b)} of site-based sparse Markovian GP regression. The kernel is Mat\'ern-\nicefrac{5}{2}, so inducing states (\protect\tikz[baseline=-2pt]{\protect\node[circle,fill=black,inner sep=0,minimum size=5pt] at (0,0){};\protect\draw[ultra thick,black](-4pt,-4pt)--(4pt,4pt);}) contain higher-order derivative information, making them more descriptive than conventional inducing function evaluations with the same number of inducing points. To achieve the same approximation accuracy, the conventional method would require the use of more function evaluations as inducing variables.
  }
  \label{fig:teaser}
  \vspace*{-1em}
\end{figure}

GPs with uni-dimensional input are especially well-suited to modeling time series and spatio-temporal data.
In this setting, the versatile class of {\em Markovian GPs} provides great computational advantages. 
These are GPs that can be rewritten in a stochastic differential equation (SDE) form \citep{sarkka2019applied} and, when marginalized to a discrete set of $N$ ordered input locations, induce a sparse precision structure which enables efficient inference algorithms with linear time computational complexity, $\bigO(N)$, as opposed to the classic cubic time scaling, $\bigO(N^3)$, usually associated with such models.

\emph{Sparse GPs} \citep{Quinonero-Candela+Rasmussen:2005, snelson2006sparse} are an alternative method for dealing with the computational intractability of GPs for large data sets, which exploit redundancy in the $N$ data points to summarise the underlying function via a smaller set of $M$ inducing points. This approach typically leads to inference algorithms with computational complexity $\bigO(NM^2+M^3)$ \citep{hensman2013gaussian}. Whilst conventional sparse GPs have been successfully applied in many domains, they are not naturally suited to time series, since the number of inducing points must grow in line with the number of time steps in order to describe all the variation in the data, which is prohibitive since their computational scaling is cubic in $M$.

For both of these schemes, methods have been devised to tackle intractable inference when using non-Gaussian likelihoods, making GPs applicable to large data sets where they provide principled uncertainty quantification and out-of-sample generalisation. These include variational inference \citep[VI,][]{titsias2009variational, hensman2013gaussian, durrande2019banded} and expectation propagation \citep[EP,][]{bui2017unifying, wilkinson2020state}.

Recently, sparsity and Markovianity have been combined under the variational inference framework to exploit the benefits of both approaches in a method named \emph{doubly sparse variational GPs} \citep[\ssvgp,][]{adam2020doubly}, where the term \emph{doubly sparse} comes from the fact that the method exploits the \emph{sparse} precision matrix of states of a Markovian GP marginalized to a finite set of \emph{sparse} inducing time points. This approach scales linearly in $(M+N)$, with $M$ sequential computations and $N$ independent ones, making it well suited to time series (see \cref{fig:teaser} for an illustration of the approach).

We generalise the doubly sparse approach by deriving \emph{site-based} approximate inference algorithms for sparse Markovian GPs. These methods contrast the existing \ssvgp method by parametrising the global approximate posterior via a set of \emph{local} contributions from the data, in the same vein as EP.
We derive four novel approximate inference algorithms based on this approach. These amount to doubly sparse extensions to conjugate-computation variational inference \citep[CVI,][]{khan2017conjugate}, power-expectation propagation \citep[PEP,][]{minka2004power}, posterior linearisation \citep[PL,][]{garcia2016iterated}, and the Extended Kalman smoother \citep[EKS,][]{bell1994iterated}. We present these algorithms alongside the existing \ssvgp approach, providing an overview of methods for inference in non-conjugate GP time series.

In our site-based algorithms, we leverage the idea of \emph{site tying} \citep{li2015stochastic} in a principled way to reduce the storage requirement of the algorithm from $\mathcal{O}(N)$ to $\mathcal{O}(Md^2)$, where $d$ is the dimensionality of the state space representation of the Markovian GP, which is extremely efficient when $M \ll N$.
We examine the properties of our new algorithms, and make detailed comparisons on multiple large time series data sets. In addition, we show how the methods can be applied to spatio-temporal data, where spatial inducing points are tracked over time by temporal processes which are summarised via a reduced number of time steps. 

An efficient JAX \citep{jax2018} implementation of all site-based methods is provided at \url{https://github.com/AaltoML/Newt}.

\section{BACKGROUND}
\label{sec:background}
Gaussian processes describe distributions over functions by stating that function evaluations at any finite collection of inputs are jointly Gaussian distributed. Given data comprising input-output pairs ${\{\vx_n, y_n\}_{n=1}^N\ \in (\XX, \RR)^N}$, they are characterised completely by mean function $\mu(\vx)$ and covariance function $\kappa(\vx, \vx')$. A GP prior over a function, $f$, and the corresponding likelihood model for the observations, $y_n$, are written,
\vspace*{-0.5em}
\begin{equation}
    \vspace*{-0.5em}
    \hspace*{-0.8em} f(\vx) \sim \GP(\mu(\vx), \kappa(\vx, \vx')), \,\,\, \vy \mid \vf \sim \prod_{n=1}^N p(y_n \mid f_n),
\end{equation}
where $\vf=f(\vX)$ and $f_n=f(\vx_n)$. If the likelihood is Gaussian, then the posterior, $p(\vf \mid \vy)$, can be computed in closed form, at a computational cost $\bigO(N^3)$. However, we are interested in the general, non-conjugate, case which necessitates approximate inference methods.

\subsection{Markovian Gaussian Processes}

Markovian Gaussian processes are GPs with one-dimensional inputs, $x\in \RR$, that have an equivalent linear time invariant (LTI) stochastic differential equation (SDE) representation with state dimension $d$:
\begin{align}
    \dot{\vs}(x) &= \vF \vs(x) + \vL \veps(x)\,, \qquad
    f(x) = \vH \vs(x)\,,
\end{align}
where $\veps(x) \in \RR^e$ is a white noise process and ${\vF \in \RR^{d\times d}}$, ${\vL \in \RR^{d\times e}}$, ${\vH \in \RR^{1\times d}}$ are the feedback, noise effect, and emission matrices. This representation supports linear time inference algorithms by explicitly performing inference over the larger set of random variables that constitute the discrete state space trajectory, $\ss = \vs(\vx)$, indexed at $\vx = [x_1,\dots, x_N]\in \RR^N$. The majority of commonly used GP kernels (on one-dimensional inputs) admit the above form \citep{sarkka2019applied}.

The solution to this LTI-SDE evaluated at $\vx$ follows a discrete-time linear system:
\begin{align}
    \vs(x_{n+1}) &= \vA_{n,{n+1}} \vs(x_n) + \vq_n, & \vq_n &\sim \NN(\bm{0}, \vQ_{n,{n+1}}),\nonumber\\
    \vs(x_0) &\sim \NN(\bm{0}, \vP_0), & \hspace*{-1em} f_n &= \vH \vs(x_n),
\end{align}
where state transitions, $\vA_{n,{n+1}} \in \RR^{d\times d}$, noise covariances, $\vQ_{n,{n+1}} \in \RR^{d\times d}$, and stationary state covariances, $\vP_0 \in \RR^{d\times d}$, can be computed analytically (see \cref{app:marginals}).
In the conjugate case, when ${p(y_n \mid f_n = \vH \vs(x_n))}$ is Gaussian, the posterior is a GP.
Its marginal statistics and the marginal likelihood, $p(\vy)$ (used for optimising the model parameters), are available in closed form and can be computed efficiently using Kalman recursions with computational scaling $\bigO(N d^3)$. For non-Gaussian likelihoods we resort to approximate inference, and various schemes have been proposed \citep{nickisch2018state, durrande2019banded, wilkinson2020state, chang2020fast}.

\subsection{Sparse Gaussian Process Approximations}
\label{sec:sparse_gps}
Sparse GPs are one of the most successful solutions to handling scalability issues and have allowed GPs to be applied to large data sets \citep[see] [for a review]{bui2017unifying}. 
The \emph{true} posterior process can be expressed as $p(f(\cdot)\mid\vy)= \int p(f(\cdot), \vf\mid\vy ) \dd\vf =  \int p(f(\cdot) \mid \vf)\,p(\vf\mid\vy)\dd\vf$. 
Here we use $f(\cdot)$ to denote all possible function evaluations (including $\vu$ and $\vf$).
This expression captures the information flow from the data $\vy$ through function evaluations $\vf=f(\vX)$. 
Sparse approximations build an \emph{approximate} posterior process of the form,
\begin{equation}
\label{eq:qf}
q(f(\cdot)) = \mint p(f(\cdot)\mid f(\vZ)=\vu) \, q(\vu)\dd\vu,
\end{equation}
where $\vZ \in \XX^M$ are referred to as \emph{pseudo-inputs}, and $q(\vu)$ can be interpreted as an approximate posterior on $\vu=f(\vZ) \in \RR^M$, i.e. $q(\vu) \approx p(\vu\mid \vy)$. This approach can be extended by conditioning the process using  deterministic functions of the process $\vu =\vphi(f)$  \citep[\eg,][]{dutordoir2020sparse, hensman2018variational}. Such approaches are referred to as \emph{inter-domain}, and the search for good inter-domain features is driven by competing demands for good approximation accuracy, tractability and scalability. This paper is based on a particular inter-domain formulation dedicated to Markovian GPs whereby the inducing variables are inducing states \citep[][see \cref{sec:inducing_states}]{adam2020doubly}. 
Given a choice of inducing variables, there are two main approaches to parametrizing the approximate posterior: a global approximation, and a local one.%

\paragraph{Global approximate posterior} One way to construct the approximate posterior is to choose $q(\vu)=\NN(\vm, \vL \vL^\top)$ to be a free-form multivariate Gaussian, whose mean $\vm \in \RR^{M}$ and Cholesky factor of the covariance $\vL \in \RR^{M \times M}$ are optimised with respect to some objective.

\paragraph{Site-based approximate posterior} An alternative, site-based approach to inference utilises the theoretical optimal form of the approximate posterior \citep{Opper+Archambeau:2009, bui2017unifying}:
\begin{equation}
\label{eq:site-based-factorisation}
    q(\vu) \propto p(\vu)\mprod_n t_n(\vu), 
\end{equation}
\ie, we assume that it factorises as a product of the prior and the (possibly unnormalized) Gaussian sites parameterized in the natural form, $t_n(\vu) = \tilde{\NN}(\vu; z_n,\vlambda_{1,n}, \vlambda_{2,n}) = z_n \exp(\vu^\top \vlambda_{1,n}  -\nicefrac{1}{2} \vu\vlambda_{2,n}\vu^\top)$, with $\vlambda_{1,n} \in \RR^{M}$ and $\vlambda_{2,n} \in \RR^{M \times M}$. These can be thought of as pseudo likelihood terms that describe the effect of the data on the posterior. This leads $q(\vu)$ to be Gaussian and its statistics can be computed in closed form.
In the algorithms we describe, the optimal sites can be shown to be rank one, \ie, $t_n(\vu) = \tilde{\NN}(\vW_n\vu; z_n, \lambda_{1,n}, \lambda_{2,n})$, where $\vW_n$ is the conditional projection matrix: $\EE_p[f_n\mid\vu] = \vW_n \vu$. The sites can be updated either via gradient-based methods or via iterative deterministic algorithms.

\subsection{Variational Inference}

\paragraph{Global VI (SVGP)}
In the global variational approach to sparse GP inference \citep[VI,][]{titsias2009variational}, one attempts to directly learn $q(\vu)$ by minimizing the KL divergence between the approximate posterior $q(f)$, \cref{eq:qf}, and the true posterior, $\vDelta=\text{KL}[q(f) \,\|\, p(f \mid \vy)]$, or equivalently by maximizing the variational objective, also called the evidence lower bound (ELBO),
\begin{align} \label{eq:var-obj}
    \cL(q) = \EE_q \log p(\vy \mid \vf) - \text{KL}[q(\vu) \,\|\, p(\vu)],
\end{align}
which verifies ${\log p(\vy) - \cL(q) = \vDelta }$. This ELBO can used both for inference and learning. 
Evaluation of the KL divergence and of the expected log-likelihood terms, also called variational expectations, have respective computational costs of $\bigO(M^3)$ and $\bigO(NM^2)$, leading to overall computational complexity $\bigO(M^3 + NM^2)$ \citep{hensman2013gaussian}. %

\paragraph{Local VI (CVI)} \label{sec:cvi} 
Conjugate-compuation variational inference \citep[CVI,][]{khan2017conjugate} uses a mirror descent algorithm to derive a site-based algorithm that is equivalent to performing VI with natural-gradients \citep{salimbeni2018natural}. To the best of our knowledge, CVI has not yet been applied to sparse GPs. To do so, the generative model must be split into a conjugate part (\ie, the prior $p(\vu)$), and a non-conjugate part which gathers the remaining terms of the likelihood $p(\vy\mid\vf)$ and the conditional prior $p(\vf\mid\vu)$:
\begin{equation}
    p(\vf, \vu, \vy) = \underbrace{p(\vu)}_{p_c(\vu)}\,\underbrace{p(\vf\mid\vu)\,p(\vy\mid\vf)}_{p_{nc}(\vf, \vu)}.
\end{equation}
CVI approximates the non-conjugate part using Gaussian sites with the sufficient statistics of $p(\vu)$: $\tilde{p}_{nc}(\vf, \vu) \approx p(\vf\mid\vu) t(\vu)$, where $t(\vu)=\prod_{n=1}^N t_n(\vu)$, which turns out to be the same parametrisation as used in EP.

Letting $\vLambda$ and $\vlambda$ be the natural parameters of the prior $p(\vu)$ and sites $t(\vu)$ respectively, the natural parameters of $q(\vu)$ are $\vLambda + \vlambda$.
One can show that a natural gradient step on the variational parameters $\vlambda$ amounts to:%
\begin{gather}\label{eq:cvi_update}
\begin{aligned}
    \vg &= \nabla_{\vmu} \EE_{q(\vu)}\EE_{p(\vf \mid \vu)} \log p(\vy\mid\vf), \\
    \vlambda^{(k+1)} &= (1-\rho) \, \vlambda^{(k)} + \rho \, \vg, 
\end{aligned}
\end{gather}
where $\vmu$ are the expectation parameters of the posterior $q(\vu)$, $k$ is the training iteration, and $\rho$ is the step size. %
It should be noted that CVI is equivalent to SVGP with natural gradients as in \citet{salimbeni2018natural}. 
A natural gradient step in the SVGP approach requires switching between the natural and moment parameterisations of the global variational distribution $q(\vu)$ (and the gradients of these operations). This is more computationally costly and prone to numerical errors than the CVI derivation.%

\subsection{Expectation Propagation}
\label{sec:ep}
The sparse variant of expectation propagation \citep[EP,][]{minka2001expectation, bui2017unifying} also uses a site-based approach, with posteriors $q(f)$ and $q(\vu)$ defined as in \cref{eq:qf,eq:site-based-factorisation} respectively. 
The EP algorithm aims to \emph{globally} minimise the forward KL divergence, $\text{KL}[p(f \mid \vy) \,\|\, q(f)]$, but since this is intractable it instead updates each site separately in an iterative fashion by minimising \emph{local} KL divergences, $t_n^{\textrm{new}}(\vu)=$
\begin{equation} \label{eq:EP-KL}
     \underset{t_n^*(\vu)}{\mathrm{arg}\min}\,\overline{\text{KL}}\left[q(f(\cdot), \vu) \frac{p(y_n\mid f_n)}{t_n^{\textrm{old}}(\vu)} \,{\Big\|}\, q(f(\cdot),\vu) \frac{t_n^*(\vu)}{t_n^{\textrm{old}}(\vu)}\right], 
\end{equation}
where $\overline{\text{KL}}$ represents the KL divergence for unnormalised distributions. This is equivalent to matching the first two moments between the approximate joint (right) and the approximate joint in which one site is replaced with the true likelihood term (left). In other words, the local site $t_n(\vu)$ is optimized in the context of the leave-one-site-out posterior, $q(f(\cdot),\vu) / t_n^{\textrm{old}}(\vu)$.
Power expectation propagation \citep[PEP,][]{minka2004power} is a generalisation of EP that minimises the $\alpha$-divergence, ${\text{D}_{\alpha}[p(f\mid\vy)\,\|\,q(f)]}$, usually implemented by raising the likelihood and site terms in \cref{eq:EP-KL} to a power of $\alpha$.

\subsection{Global, Local, and Intermediate Approximations}

\citet{minka2001expectation} showed that PEP corresponds to variational algorithms in the limit of $\alpha \to 0$. That is, for $\alpha \to 0$, if PEP converges, then it converges to the same optima as that given by optimising \cref{eq:var-obj}. This result extends to the corresponding sparse VI and PEP algorithms \citep{bui2017unifying}.

In order to reduce the memory requirements associated with storing all the EP parameters, \emph{tied} sites were introduced in an algorithm called stochastic expectation propagation \cite[SEP,][]{li2015stochastic}. In the most extreme instantiation of SEP, each of the $N$ sites are set to correspond to a fraction of a global site $t_n(\vu) = t(\vu)^{1/N}$. Intermediate algorithms are also possible in which subsets of data points are tied together. These algorithms, also referred to as average EP \citep{dehaene2018expectation}, lead to an approximation whose memory requirement no longer scales with the number of data points.

In the opposite direction, efforts aimed at speeding up computation of the VI approximation have led to a localized (or de-globablized) variational posterior. For example, additional conditional independence assumptions between subsets of observations and subsets of the latent process have been proposed \citep{bui2014tree}, leading to factors of the variational distribution impacting the posterior
distribution locally. %

\subsection{Comparison and Performance Guarantees}

Overall, sparse power EP and VI approaches are efficient and performant. The most recent survey and comparison of these methods \citep{bui2017unifying} reports an overall slight advantage for EP in non-conjugate tasks. However, the VI algorithm is simpler, very modular, and has formed the basis of more extensions in the research community.

Many of the algorithms presented above are specific instances of broader classes of algorithms, and therefore inherit some general guarantees in terms of convergence or approximation error: {\em (i)}~in the variational setting, sparse GPs come with guarantees on the quality of the posterior approximation as the number of inducing point is increased \citep{burt2019rates}, {\em (ii)}~iterative updates of CVI algorithms will increase the ELBO and converge under mild conditions \citep[Prop.~2--3]{khan2016faster}, {\em (iii)}~recent convergence results also exist for EP \citep{dehaene2018expectation} under rather restrictive conditions. The search for guarantees for EP is an active research question.

\section{INDUCING STATES FOR MARKOVIAN GP MODELS}
\label{sec:inducing_states}
Despite their success in the large data regime, the computational complexity of the above sparse approximations still makes them unsuitable for long (or unbounded) time series because in order to accurately approximate the posterior, the number of inducing variables, $M$, needs to grow with the temporal horizon.
Crucially, the posterior prediction of a single data point depends on the entire set of inducing variables $\vu$ through the conditional $p(f(\cdot)\mid\vu)$, even those far apart in time.

In the following sections, we describe how the combination of Markovian GPs with sparse GPs, via state inducing features, further reduces the complexity of the algorithms, making them applicable to long time series.

\subsection{State Inducing Features}

A key property of the SDE formulation of Markovian GPs is that the state variables $\vs(\vx)$, obtained by marginalizing the SDE to inputs $\vx = (x_1, \dots, x_n)$, have a Markovian property, \ie, $p(\vs(x_n)\mid \vs(x_{n-1, \dots, 1})) = p(\vs(x_n) \mid \vs(x_{n-1}))$, which is another way of formulating the definition of the state $\vs(x)$ as a summary of all the information necessary to predict the future beyond $x$.
Thus, a natural choice of inducing variables for sparse inference with Markovian GPs is state evaluations, $\vu=\vs(\vZ)$, indexed at $M$ pseudo input locations $\vZ=(z_1, \dots, z_M)$. 

This leads to the conditional $f\mid \vu$ being local, \ie, if ${z_m \leq x_n < z_{m+1}}$, and noting $\vv_{m(n)}=[\vu_m, \vu_{m+1}]$, then ${p(f_n\mid\vu) = p(f_n\mid\vv_{m(n)})} =\NN(f_n\mid \vW_n \vv_{m(n)}, \nu_n)$. 
The conditional is available in closed form via the statistics of the prior transitions, $p(\vs_n\mid\vu_m)$ and $p(\vu_{m+1}\mid\vs_n)$, and from the emission matrix $\vH$ (see \cref{app:conditionals}). This makes marginal prediction ${q(f_n)}$, \cref{eq:qf}, cheap to evaluate since it only depends on the \emph{local} marginal posterior $q(\vv_{m(n)})$.

It should be noted that although the number of inducing points $\vZ$ is $M$, the number of inducing variables contained in $\vu$ is $Md$, where $d$ is the state dimension. Indeed each inducing state $\vs(z_m)$ contains more information than a single inducing function evaluation $f(z_m) \in \vs(z_m)$. In practice, fewer inducing inputs are needed when using inducing states than when using the classic inducing function evaluations \citep[see \cref{fig:teaser} and][]{adam2020doubly}.

\subsection{`Doubly Sparse' Variational Inference}
\label{sec:s2vgp}
The \ssvgp algorithm \citep{adam2020doubly} parameterizes an approximate posterior over the inducing states, $\vu = \vs(z_1), \dots, \vs(z_M)$, as a linear Gaussian state space model: $q(\vu)= q(\vu_1)\prod_m q(\vu_{m+1}\mid\vu_{m})$. This shared chain structure with the marginal prior $p(\vu)$ is optimal. 
The ELBO can be written as the sums:
\begin{align}
    \cL(q) &=    \msum_n \EE_{q(f_n)} \log p(y_n \mid f_n) \nonumber \\
    &-  \msum_m \text{KL}\left[q(\vu_{m+1} \mid \vu_{m}) \,\|\, p(\vu_{m+1}\mid \vu_{m})\right].
\end{align}
The marginal posterior predictions $q(f_n)$ can be evaluated independently given the pairwise marginal on the inducing states $q(\vv_m)$. These can be computed in linear time with chain length $M$ using classic Kalman filtering algorithms \citep{sarkka2013bayesian} or linear algebra routines dedicated to banded matrices \citep{durrande2019banded}.
Evaluation of the KL divergence and the variational expectations have respective computational costs of $\bigO(Md^3)$ and $\bigO(Nd^3)$, leading to an overall computational complexity of $\bigO((N + M)d^3)$.
More details on this algorithm are given in \cref{app:ssvgp}.

\section{SITE-BASED SPARSE MARKOVIAN GPs}
\label{sec:site-based-markov}

In \cref{sec:sparse_gps}, we reviewed three common algorithms used to perform approximate inference given a sparse formulation of GPs: VI, CVI and PEP. In the special case of sparse Markovian GPs using inducing states, only the VI formulation  \citep[\cref{sec:s2vgp} and][]{adam2020doubly} has been explored. 
CVI and PEP operate on the precision of the approximating distribution $q(\vu)$ and turn out to be ideally suited to the Markovian setting where this precision is sparse.
In the following sections we describe how to adapt these algorithms to this setting and show how these methods inherit the favourable properties of their parents.
We call these algorithms \sscvi and \sspep. We then go on to show that the doubly sparse approach is even more general, deriving the equivalent algorithms for the classical nonlinear Kalman smoothers, \sspl and \sseks.

Using state inducing features, the optimal sites for each data point $x_n$ are only functions of the neighbouring states $\vv_{m(n)}$ due the local structure of the conditional $f_n\mid \vu=f_n\mid \vv_{m(n)}$. The approximating distribution $q(\vu)=p(\vu)\prod_n t_n(\vv_{m(n)})$ thus still has a chain structure which we exploit to efficiently compute the marginal statistics $q(\vv_m)$ via, \eg, filtering methods (see \cref{app:inference} for details). 

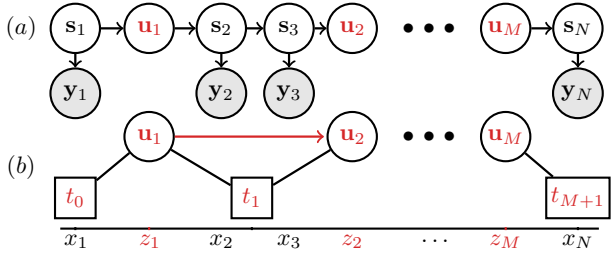
\begin{figure}[t]
\def\nang{25}

\newlength{\ndia}
\setlength{\ndia}{.25cm}
\newlength{\ndi}
\setlength{\ndi}{.6cm}

\tikzset{
  graphmod ss/.style={->,shorten >=1pt,line width=0.12\ndia, draw=red},
  graphmod ssk/.style={->,shorten >=0pt,line width=0.12\ndia, draw=black},
  graphmod sf/.style={-,shorten >=0pt,line width=0.12\ndia, draw=black},
  graphmod nbr/.style={-,line width=0.06\ndia},
  graphmod nbr red/.style={-,line width=0.06\ndia, draw=red},
  func/.style={draw,circle,inner sep=0pt,text width=7mm,text centered},
  dummy/.style={draw=none,circle,inner sep=0pt,text width=7mm,text centered},
  state/.style={draw,circle,inner sep=0pt,text width=7mm,text centered, red,draw=black},
  site/.style={draw,rectangle,minimum height=6mm, minimum width=6mm, inner sep=2pt,text centered, red,draw=black},
  random white/.style={draw,circle,inner sep=0pt,text width=7mm,text centered,draw=white},
  text/.style={draw,circle,inner sep=0pt,text width=20mm,text centered,draw=white},
}

\def\tikzdots#1--#2;{
  \path #1 -- #2
     node[pos=0.3]{$\bullet$}
     node[pos=0.5]{$\bullet$}
     node[pos=0.7]{$\bullet$};
}

\begin{center}
\begin{tikzpicture}[thick,scale=0.8, every node/.style={scale=0.9}]

\node (x1) [dummy] {$x_1$};
\node (z1) [dummy, right = .5\ndi of x1, red] {$z_1$};
\node (x2) [dummy, right = 2.1\ndi of x1] {$x_2$};
\node (x2p5) [dummy, right = 2.8\ndi of x1] {};
\node (x3) [dummy, right = 3.6\ndi of x1] {$x_3$};
\node (z2) [dummy, right = 5.\ndi of x1, red] {$z_2$};
\node (x4) [dummy,  right = 6.9\ndi of x1] {$\dots$};
\node (z3) [dummy,  right = 8.4\ndi of x1, red] {$z_M$};
\node (xN) [dummy,  right = 10\ndi of x1] {$x_N$};

\newcommand \yf {1.2\ndi};
\newcommand \yu {2.9\ndi};
\newcommand \yline {.35\ndi};
\newcommand \ydata {4.1\ndi};
\newcommand \yprior {5.9\ndi};

\path let \p0 = (x1) in node[site] (f1) at (\x0,\yf ) {$t_0$};
\path let \p0 = (x2p5) in node[site] (f2) at (\x0,\yf ) {$t_1$};
\path let \p0 = (xN) in node[site] (fN) at (\x0,\yf ) {$t_{M+1}$};
\path let \p0 = (z1) in node[state] (s1) at (\x0,\yu) {$\vec u_1$};
\path let \p0 = (z2) in node[state] (s2) at (\x0,\yu) {$\vec u_2$};
\path let \p0 = (z3) in node[state] (s3) at (\x0,\yu) {$\vec u_M$};

\draw (-\ndia,\yline) -- (14.5\ndi,\yline);

\draw[thick] let \p{f}=(f1) in (\x{f},\yline-.1\ndia)--(\x{f},\yline+.1\ndia);
\draw[thick] let \p{f}=(f2) in (\x{f},\yline-.1\ndia)--(\x{f},\yline+.1\ndia);
\draw[thick] let \p{f}=(fN) in (\x{f},\yline-.1\ndia)--(\x{f},\yline+.1\ndia);

\draw[thick,red] let \p{f}=(s1) in (\x{f},\yline-.1\ndia)--(\x{f},\yline+.1\ndia);
\draw[thick,red] let \p{f}=(s3) in (\x{f},\yline-.1\ndia)--(\x{f},\yline+.1\ndia);

\newcommand \opac {1.};

\newcommand \dso {.5};

\draw [graphmod sf] (s1) -- (f2) node[left,pos=0.5]{} ;
\draw [graphmod sf] (s2) -- (f2) node[left,pos=0.5]{} ;

\draw [graphmod sf, opacity=\opac] (s1) -- (f1) node[left,pos=0.5]{} ;
\draw [graphmod sf, opacity=\opac] (s3) -- (fN) node[left,pos=0.5]{} ;

\draw [graphmod ss] (s1) -- (s2) node[left,pos=0.5]{} ;

\path let \p0 = (x1) in node[dummy] (a) at (\x0-1.5\ndi, \yprior ) {$(a)$};
\path let \p0 = (x1) in node[dummy] (b) at (\x0-1.5\ndi, .5*\yu+.5*\yf){$(b)$};

\path let \p0 = (x1) in node[func] (s1p) at (\x0,\yprior ) {$\vec s_1$};
\path let \p0 = (x2) in node[func] (s2p) at (\x0,\yprior ) {$\vec s_2$};
\path let \p0 = (x3) in node[func] (s3p) at (\x0,\yprior ) {$\vec s_3$};
\path let \p0 = (xN) in node[func] (sNp) at (\x0,\yprior ) {$\vec s_N$};
\path let \p0 = (z1) in node[state] (u1p) at (\x0,\yprior) {$\vec u_1$};
\path let \p0 = (z2) in node[state] (u2p) at (\x0,\yprior) {$\vec u_2$};
\path let \p0 = (z3) in node[state] (u3p) at (\x0,\yprior) {$\vec u_M$};

\draw [graphmod ssk, black] (s1p) -- (u1p) node[left,pos=0.5]{} ;
\draw [graphmod ssk, black] (u1p) -- (s2p) node[left,pos=0.5]{} ;
\draw [graphmod ssk, black] (s2p) -- (s3p) node[left,pos=0.5]{} ;
\draw [graphmod ssk, black] (s3p) -- (u2p) node[left,pos=0.5]{} ;
\draw [graphmod ssk, black] (u3p) -- (sNp) node[left,pos=0.5]{} ;

\path let \p0 = (x1) in node[func,fill=black!10] (y1) at (\x0,\ydata ) {$\vec y_1$};
\path let \p0 = (x2) in node[func,fill=black!10] (y2) at (\x0,\ydata ) {$\vec y_2$};
\path let \p0 = (x3) in node[func,fill=black!10] (y3) at (\x0,\ydata ) {$\vec y_3$};
\path let \p0 = (xN) in node[func,fill=black!10] (yN) at (\x0,\ydata ) {$\vec y_N$};

\draw [graphmod ssk, black] (s1p) -- (y1) node[left,pos=0.5]{} ;
\draw [graphmod ssk, black] (s2p) -- (y2) node[left,pos=0.5]{} ;
\draw [graphmod ssk, black] (s3p) -- (y3) node[left,pos=0.5]{} ;
\draw [graphmod ssk, black] (sNp) -- (yN) node[left,pos=0.5]{} ;

\tikzdots (u2p) -- (u3p); 
\tikzdots (s2) -- (s3);

\end{tikzpicture} 
\end{center}
\caption{\textbf{(a)}~Directed graphical model representing the joint prior over states indexed at $x_n$, associated data points $y_n$, and inducing states $\vec{u}_m$ indexed at $z_m$. \textbf{(b)}~Graphical model for shared site based approximate posterior, with local sites $t_{m}(\vec{v}_m)$ represented as factor graph (undirected black edges). The prior over inducing states $p(\vu)$ is a state space model represented as a directed graphical model (red arrows).}
\label{fig:graph_model}
\end{figure}

A consequence of the locality of the sites is that all data points who lie in the same time segment between consecutive inducing inputs $[z_m, z_{m+1}]$ share the same support, $\vv_m$. This provides a natural way to tie these sites together per segment $t_m(\vv_m) = \prod_{n \in \mathcal{M}_m} t_n(\vv_m)$, where $\mathcal{M}_m$ represents the indices to the data whose inputs fall in $[z_m, z_{m+1}]$. We adopt this approach, reducing our algorithms' memory requirements to $\mathcal{O}(Md^2)$, which is equivalent to \ssvgp. A graphical representation of the approach is depicted in \cref{fig:graph_model}

\begin{figure*}[t!]
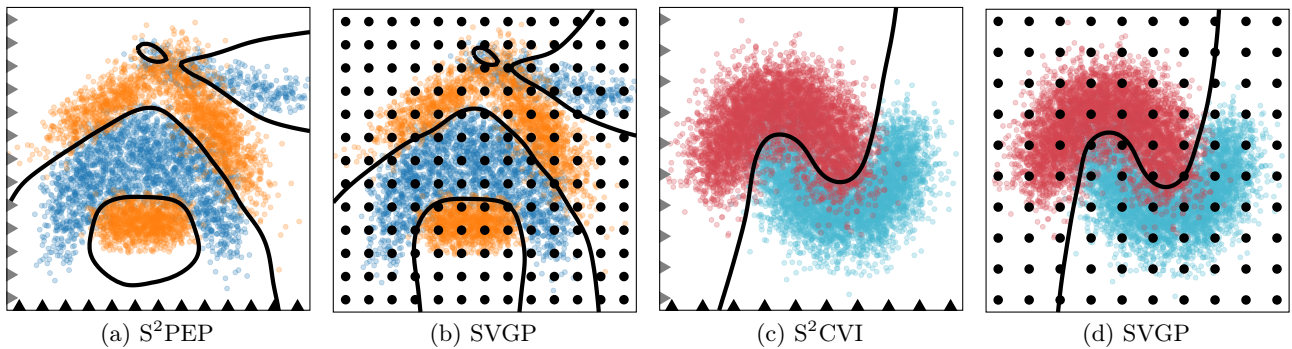

  \centering
  \pgfplotsset{yticklabel style={rotate=90}, ylabel style={yshift=0pt},scale only axis,axis on top,clip=true,clip marker paths=true}
  \setlength{\figurewidth}{.235\textwidth}
  \setlength{\figureheight}{\figurewidth}
  \begin{subfigure}[b]{.5\columnwidth}
    \centering
    \input{./graphs/banana-s2pep.tex}
    \vspace{-0.5cm}
    \caption{\sspep}
    \label{fig:sspep-banana}  
  \end{subfigure}
  \hfill  
  \begin{subfigure}[b]{.5\columnwidth}
    \centering
    \input{./graphs/banana-svgp.tex}
    \vspace{-0.5cm}
    \caption{SVGP}
    \label{fig:svgp-banana}  
  \end{subfigure}
  \hfill
  \begin{subfigure}[b]{.5\columnwidth}
    \centering
    \input{./graphs/moon-s2cvi.tex}
    \vspace{-0.5cm}
    \caption{\sscvi}
    \label{fig:sscvi-moon}  
  \end{subfigure}
  \hfill
  \begin{subfigure}[b]{.5\columnwidth}
    \centering
    \input{./graphs/moon-svgp.tex}
    \vspace{-0.5cm}
    \caption{SVGP}
    \label{fig:svgp-moon}  
  \end{subfigure}\\[-.2cm]
  \caption{2D classification tasks. For the doubly sparse methods, {(a)} and {(c)}, inducing points are placed separately in the sequential dimension ($x$-axis, `time', black triangles) and the `spatial' dimension ($y$-axis, grey triangles).}
  \label{fig:banana}
  \vspace*{-0.25cm}
\end{figure*}

\subsection{Doubly Sparse CVI (\sscvi)} \label{sec:sscvi}
In the sparse state space formulation, the prior on the inducing states $p(\vu)$ has sufficient statistics ${\phi(\vu)=[{(\vu_k, \vu_k \vu_k^\top)}^M_{k=1}, (\vu_{k+1} \vu_k^\top)_{k=1}^{M-1}]}$ whose associated second-order natural parameters are the block-tridiagonal entries of the sparse precision matrix of $p(\vu)$. 
The non-tied formulation of CVI would introduce $N$ sites, each dependent only on their nearest inducing states $t_n(\vv_{m(n)})$. Indeed, because $p(f_n\mid\vu)= p(f_n\mid\vv_{m(n)})$, the site update $\vg_n = \nabla_{\vmu} \EE_{q(f_n)} \log p(y_n\mid f_n)$ in \cref{eq:cvi_update} is only non-zero for the natural parameter associated to the sufficient statistics of site $t_n$, \ie, $[\vv_{m(n)}, \vv_{m(n)}\vv_{m(n)}^\top]$.

Here we use tied sites, parameterising $M+1$ Gaussian sites $t_m(\vv_m)$ with sufficient statistics $[\vv_m, \vv_m \vv_m^\top]$ in their natural form.
The edge cases are smaller sites over the first and last inducing states $t_0(\vu_0=\vs(-\infty))$ and $t_{M+1}(\vu_{M+1}=\vs(\infty))$. %

Unlike our presentation in \cref{sec:cvi}, the sites are local, only depending on the states that directly neighbour them. The update rule is the same as in \cref{eq:cvi_update}, but now using the fact that $p(f_n\mid\vu)= p(f_n\mid\vv_{m(n)})$, and that gradient $\vg_n$  only contributes to site $t_{m(n)}(\vv_{m(n)})$.

The updates can be written in terms of gradients of the variational expectations with respect to the mean and variance of the posterior marginal via the chain rule,
\begin{gather}
\begin{aligned} \label{eq:param-update-cvi}
    &\vg_{n,2} = \vW_n^\top \diff{\mathcal{L}_n}{\Sigma_n} \vW_n, \\
    &\vg_{n,1} = \vW_n^\top \diff{\mathcal{L}_n}{\mu_n} - 2 \vW_n^\top \diff{\mathcal{L}_n}{\Sigma_n} \vW_n \vmu_{m(n)},
\end{aligned}
\end{gather}
where $\mathcal{L}_n=\EE_{q(f_n)} \log p(y_n\mid f_n)$ and $\mu_n=\vW_n\vmu_{m(n)}$, $\Sigma_n=\vW_n \vSigma_{m(n)} \vW_n^\top + \nu_n$ are the moments of $q(f_n)$.
This algorithm is equivalent to the natural gradient approach presented in \citet{adam2020doubly}. Its practical implementation is however simpler and less costly since there is no need to compute the KL term of the ELBO to perform the update of the variational parameters. As a result it is also more numerically stable.
The full details of the algorithm are given in \cref{app:sscvi}. %

\subsection{Doubly Sparse Power Expectation Propagation (\sspep)}\label{sec:sspep}

The \sspep algorithm approximates the joint distribution over the states and the observations,
\begin{align}
    \!\!\!\!p(\vs(\cdot) , \vy) & =  p(\vu)\,p(\vs(\cdot)\mid\vu)\mprod_n p(y_n\mid\vs_n) \nonumber\\
    & \approx p(\vu)\,p(\vs(\cdot)\mid\vu)\mprod_m t_m(\vv_m) = q(\vs(\cdot)),  
\end{align}
where $t_m(\vv_m)$ is the tied site for all $x\in \XX_m$, where $\XX_m = \{x \in \vx\mid {z_m\leq x< z_{m+1}\}}$. We can obtain the site for a single data point $x_n$ as $t_n(\vv_{m(n)})=t^{\nicefrac{1}{N_{m(n)}}}_{m(n)}(\vv_{m(n)})$, where $N_{m(n)}=|\XX_m|$ is the total number of data points in the neighbourhood.
We now outline the PEP steps for updating the sites.%

\paragraph{Cavity computation} The leave-one-site-out posterior, \ie, the cavity, for a given data point $y_n$ is determined first by computing the approximate posterior over the state with a fraction $\nicefrac{\alpha}{N_{m(n)}}$ of the local site removed, 
\begin{align} %
    q^\textrm{cavity}(\vv_{m(n)}) &\propto q(\vv_{m(n)}) \big/ t_{m(n)}^{\nicefrac{\alpha}{N_{m(n)}}}(\vv_{m(n)}) \nonumber \\
    &= \NN(\vv_{m(n)} \mid \vmu_{m(n)}, \vSigma_{m(n)}). 
\end{align}
The cavity on function evaluation $f_n$ is obtained by marginalizing the joint cavity over $f_n$ and $ \vv_{m(n)}$:
\begin{align} 
\label{eq:sspep-cavity}
    & q^\textrm{cavity}(f_n) = \mint  q^\textrm{cavity}(f_n, \vv_{m(n)}) \dd\vv_{m(n)}  \nonumber \\
    & = \mint p(f_n \mid \vv_{m(n)})q^\textrm{cavity}(\vv_{m(n)}) \dd\vv_{m(n)} \\
    &= \NN(f_n \mid \mu_n=\vW_n\vmu_{m(n)}, \vW_n\vSigma_{m(n)} \vW_n^\top + \nu_n). \nonumber
\end{align}

\paragraph{Moment matching} We next compute the moments of the so-called \emph{tilted} distribution, \ie, the cavity combined with (a fraction of) the true likelihood function. As discussed in \citet{bui2017unifying} and \citet{seeger2005expectation}, the required moments can be conveniently obtained via the derivatives of the log-normaliser of the tilted distribution, $\log Z_n = \log \EE_{q^\textrm{cavity}}[p^{\alpha}(y_n \mid f_n)]$, with respect to the cavity mean. Doing so provides the new marginal posterior moments ${q(\vv_{m(n)})=\NN(\vv_{m(n)}\mid \vmu^\textrm{post}_{{m(n)}}, \vSigma^\textrm{post}_{{m(n)}})}$:
\begin{gather}
\begin{aligned} \label{eq:moment-match}
    \vmu^\textrm{post}_{{m(n)}} &= \vmu_{m(n)} + \vW_{n} \diff{\log Z_n}{\mu_n },\\
    \quad \vSigma^\textrm{post}_{{m(n)}} &= \vSigma_{m(n)} + \vW_{n} \diffII{\log Z_n}{\mu_n }\vW_{n}^\top . %
\end{aligned}
\end{gather}
For Gaussian likelihoods, the above derivatives are available in closed form, whilst for non-conjugate models we must resort to numerical integration. Given the new marginal posterior, we can finally compute the new tied site by removing the cavity from the posterior and combining it with a fraction of the old site (representing the other data points in the neighbourhood):
\begin{equation}
    t^\textrm{new}_{m(n)}(\vv_{m(n)}) = \left[t^\textrm{old}_{m(n)}(\vv_{m(n)}) \right]^{1-\frac{\alpha}{N_{m}}} \left[ \frac{q(\vv_{m(n)})}{q^\textrm{cavity}(\vv_{m(n)})} \right].%
\end{equation}
\vspace{-0.5cm}

\begin{figure*}[t!]
	\centering\scriptsize
	\tikzstyle{every picture}+=[remember picture]
	\pgfplotsset{scaled ticks=false,yticklabel style={rotate=90}, ylabel style={yshift=0pt}, xlabel style={yshift=3pt}, scale only axis,axis on top,clip=true,clip marker paths=true, xtick align=inside}
	\pgfplotsset{legend style={inner xsep=1pt, inner ysep=1pt, row sep=0pt},legend style={at={(0.98,0.98)},anchor=north east},legend style={rounded corners=1pt},legend style={fill=white, fill opacity=0.8, draw opacity=1, text opacity=1, draw=white!80!black},legend style={nodes={scale=0.8, transform shape}}}
	\setlength{\figurewidth}{.28\textwidth}
	\setlength{\figureheight}{0.9\figurewidth}
	\begin{subfigure}[b]{.32\textwidth}
	    \centering
\begin{tikzpicture}

\definecolor{color0}{rgb}{0.12156862745098,0.466666666666667,0.705882352941177}
\definecolor{color1}{rgb}{1,0.498039215686275,0.0549019607843137}
\definecolor{color2}{rgb}{0.172549019607843,0.627450980392157,0.172549019607843}
\definecolor{color3}{rgb}{0.83921568627451,0.152941176470588,0.156862745098039}
\definecolor{color4}{rgb}{0.580392156862745,0.403921568627451,0.741176470588235}
\definecolor{color5}{rgb}{0.549019607843137,0.337254901960784,0.294117647058824}

\begin{axis}[
height=\figureheight,
legend cell align={left},
legend style={fill opacity=0.8, draw opacity=1, text opacity=1, draw=white!80!black},
tick pos=left,
width=\figurewidth,
x grid style={white!69.0196078431373!black},
xlabel={Number of inducing inputs, \(\displaystyle M\)},
xmin=2, xmax=16,
xtick style={color=black},
y grid style={white!69.0196078431373!black},
ymin=950.393543163107, ymax=3379.21142711606,
ytick style={color=black}
]
\path [fill=color0, fill opacity=0.05]
(axis cs:2,3201.5948550478)
--(axis cs:2,3190.65319630071)
--(axis cs:4,2725.28827149358)
--(axis cs:6,2261.730313816)
--(axis cs:8,1643.81103504747)
--(axis cs:10,1398.28534315516)
--(axis cs:12,1217.03638819982)
--(axis cs:14,1101.74092332059)
--(axis cs:16,1069.8910005984)
--(axis cs:18,1065.23546488942)
--(axis cs:20,1064.76220065445)
--(axis cs:20,1075.29257503468)
--(axis cs:20,1075.29257503468)
--(axis cs:18,1075.96637758409)
--(axis cs:16,1081.28215294099)
--(axis cs:14,1113.31377361501)
--(axis cs:12,1228.98262797639)
--(axis cs:10,1411.7095029172)
--(axis cs:8,1659.56865010084)
--(axis cs:6,2286.33005079979)
--(axis cs:4,2750.95680436501)
--(axis cs:2,3201.5948550478)
--cycle;

\path [fill=color1, fill opacity=0.05]
(axis cs:2,3133.1275409849)
--(axis cs:2,3126.39065134774)
--(axis cs:4,2477.31271512588)
--(axis cs:6,2074.22049600659)
--(axis cs:8,1676.67886082656)
--(axis cs:10,1379.14016286034)
--(axis cs:12,1209.36319548053)
--(axis cs:14,1095.16254857011)
--(axis cs:16,1064.57829964953)
--(axis cs:18,1061.24757337226)
--(axis cs:20,1060.91541742995)
--(axis cs:20,1071.72396956294)
--(axis cs:20,1071.72396956294)
--(axis cs:18,1072.10689391393)
--(axis cs:16,1075.97918079642)
--(axis cs:14,1106.71447925568)
--(axis cs:12,1220.60983371413)
--(axis cs:10,1392.81237550818)
--(axis cs:8,1697.74825717069)
--(axis cs:6,2092.32162472203)
--(axis cs:4,2498.50294411306)
--(axis cs:2,3133.1275409849)
--cycle;

\path [fill=color2, fill opacity=0.05]
(axis cs:2,3179.58748117915)
--(axis cs:2,3168.04107495294)
--(axis cs:4,2615.13325746976)
--(axis cs:6,2188.98084082079)
--(axis cs:8,1617.00644066462)
--(axis cs:10,1378.24851165477)
--(axis cs:12,1208.24583363326)
--(axis cs:14,1095.22076547049)
--(axis cs:16,1064.60825844965)
--(axis cs:18,1061.21941919712)
--(axis cs:20,1060.79435607006)
--(axis cs:20,1071.46145993018)
--(axis cs:20,1071.46145993018)
--(axis cs:18,1072.08466170548)
--(axis cs:16,1076.01649950102)
--(axis cs:14,1106.78565757725)
--(axis cs:12,1219.73205009244)
--(axis cs:10,1391.61383139026)
--(axis cs:8,1633.02182055918)
--(axis cs:6,2209.9509048199)
--(axis cs:4,2637.71249473957)
--(axis cs:2,3179.58748117915)
--cycle;

\path [fill=color3, fill opacity=0.05]
(axis cs:2,3259.32976634847)
--(axis cs:2,3254.90759333383)
--(axis cs:4,2868.83084804807)
--(axis cs:6,2356.80655951406)
--(axis cs:8,1772.62598551629)
--(axis cs:10,1464.6462721894)
--(axis cs:12,1242.4058583373)
--(axis cs:14,1108.34904089476)
--(axis cs:16,1072.84825435723)
--(axis cs:18,1067.5856284727)
--(axis cs:20,1066.32340531208)
--(axis cs:20,1077.08801324882)
--(axis cs:20,1077.08801324882)
--(axis cs:18,1078.52143857554)
--(axis cs:16,1084.21019801619)
--(axis cs:14,1119.66179398834)
--(axis cs:12,1254.09864481114)
--(axis cs:10,1478.44224717273)
--(axis cs:8,1788.50130227459)
--(axis cs:6,2374.62443293181)
--(axis cs:4,2889.44299168965)
--(axis cs:2,3259.32976634847)
--cycle;

\path [fill=color4, fill opacity=0.05]
(axis cs:2,3268.7018046773)
--(axis cs:2,3264.96078064388)
--(axis cs:4,2972.99490930351)
--(axis cs:6,2456.48462850999)
--(axis cs:8,1868.59218595343)
--(axis cs:10,1532.53744341776)
--(axis cs:12,1273.2899161608)
--(axis cs:14,1116.68865542345)
--(axis cs:16,1077.21421005662)
--(axis cs:18,1071.41959104928)
--(axis cs:20,1074.8944731794)
--(axis cs:20,1085.87716022916)
--(axis cs:20,1085.87716022916)
--(axis cs:18,1082.52921511035)
--(axis cs:16,1088.51813968473)
--(axis cs:14,1127.79448630146)
--(axis cs:12,1285.50357381576)
--(axis cs:10,1547.3238019506)
--(axis cs:8,1885.10121354669)
--(axis cs:6,2474.34495406722)
--(axis cs:4,2992.2682969464)
--(axis cs:2,3268.7018046773)
--cycle;

\path [fill=color5, fill opacity=0.05]
(axis cs:2,3268.81061420911)
--(axis cs:2,3265.07807108853)
--(axis cs:4,2974.10532021858)
--(axis cs:6,2457.82366267815)
--(axis cs:8,1869.98594872792)
--(axis cs:10,1533.7184524682)
--(axis cs:12,1273.87791294418)
--(axis cs:14,1116.83257995998)
--(axis cs:16,1077.26670948175)
--(axis cs:18,1071.1033177229)
--(axis cs:20,1074.95820648438)
--(axis cs:20,1085.94088306806)
--(axis cs:20,1085.94088306806)
--(axis cs:18,1082.09123176125)
--(axis cs:16,1088.56935659209)
--(axis cs:14,1127.93594200279)
--(axis cs:12,1286.10388771393)
--(axis cs:10,1548.5196026642)
--(axis cs:8,1886.50711538057)
--(axis cs:6,2475.68253200341)
--(axis cs:4,2993.3557002979)
--(axis cs:2,3268.81061420911)
--cycle;

\addplot [thick, color0]
table {%
2 3196.12402567426
4 2738.12253792929
6 2274.03018230789
8 1651.68984257415
10 1404.99742303618
12 1223.0095080881
14 1107.5273484678
16 1075.58657676969
18 1070.60092123675
20 1070.02738784456
};
\addlegendentry{S$^2$EKS}
\addplot [thick, color1]
table {%
2 3129.75909616632
4 2487.90782961947
6 2083.27106036431
8 1687.21355899863
10 1385.97626918426
12 1214.98651459733
14 1100.9385139129
16 1070.27874022298
18 1066.6772336431
20 1066.31969349644
};
\addlegendentry{S$^2$PL}
\addplot [thick, color2]
table {%
2 3173.81427806605
4 2626.42287610467
6 2199.46587282034
8 1625.0141306119
10 1384.93117152251
12 1213.98894186285
14 1101.00321152387
16 1070.31237897534
18 1066.6520404513
20 1066.12790800012
};
\addlegendentry{S$^2$PEP($\alpha=1$)}
\addplot [thick, color3]
table {%
2 3257.11867984115
4 2879.13691986886
6 2365.71549622293
8 1780.56364389544
10 1471.54425968107
12 1248.25225157422
14 1114.00541744155
16 1078.52922618671
18 1073.05353352412
20 1071.70570928045
};
\addlegendentry{S$^2$PEP($\alpha=0.5$)}
\addplot [thick, color4]
table {%
2 3266.83129266059
4 2982.63160312496
6 2465.41479128861
8 1876.84669975006
10 1539.93062268418
12 1279.39674498828
14 1122.24157086246
16 1082.86617487068
18 1076.97440307982
20 1080.38581670428
};
\addlegendentry{S$^2$PEP($\alpha=0.01$)}
\addplot [thick, color5]
table {%
2 3266.94434264882
4 2983.73051025824
6 2466.75309734078
8 1878.24653205424
10 1541.1190275662
12 1279.99090032906
14 1122.38426098139
16 1082.91803303692
18 1076.59727474208
20 1080.44954477622
};
\addlegendentry{S$^2$CVI}
\end{axis}

\end{tikzpicture}
	    \vspace{-0.2cm}
	    \caption{NLML}
	\end{subfigure}
	\hfill
	\begin{subfigure}[b]{.32\textwidth}
	    \centering
\begin{tikzpicture}

\definecolor{color0}{rgb}{0.12156862745098,0.466666666666667,0.705882352941177}
\definecolor{color1}{rgb}{1,0.498039215686275,0.0549019607843137}
\definecolor{color2}{rgb}{0.172549019607843,0.627450980392157,0.172549019607843}
\definecolor{color3}{rgb}{0.83921568627451,0.152941176470588,0.156862745098039}
\definecolor{color4}{rgb}{0.580392156862745,0.403921568627451,0.741176470588235}
\definecolor{color5}{rgb}{0.549019607843137,0.337254901960784,0.294117647058824}

\begin{axis}[
height=\figureheight,
tick pos=left,
width=\figurewidth,
x grid style={white!69.0196078431373!black},
xlabel={Number of inducing inputs, \(\displaystyle M\)},
xmin=2, xmax=16,
xtick style={color=black},
y grid style={white!69.0196078431373!black},
ymin=0.177041139146669, ymax=0.703964096894227,
ytick style={color=black}
]
\path [fill=color0, fill opacity=0.05]
(axis cs:2,0.671712735505509)
--(axis cs:2,0.658464224373682)
--(axis cs:4,0.523338149475282)
--(axis cs:6,0.433200394445929)
--(axis cs:8,0.293368158723885)
--(axis cs:10,0.256770961732598)
--(axis cs:12,0.226636259136432)
--(axis cs:14,0.207978132557424)
--(axis cs:16,0.203311300466887)
--(axis cs:18,0.202831027541458)
--(axis cs:20,0.202914145546011)
--(axis cs:20,0.221442388744771)
--(axis cs:20,0.221442388744771)
--(axis cs:18,0.221476619778152)
--(axis cs:16,0.222738830619094)
--(axis cs:14,0.227013481896892)
--(axis cs:12,0.246947330999398)
--(axis cs:10,0.279856251670801)
--(axis cs:8,0.319503762123422)
--(axis cs:6,0.464038146968146)
--(axis cs:4,0.556538226375631)
--(axis cs:2,0.671712735505509)
--cycle;

\path [fill=color1, fill opacity=0.05]
(axis cs:2,0.637966936875157)
--(axis cs:2,0.618529954291892)
--(axis cs:4,0.46254475384251)
--(axis cs:6,0.401681655072669)
--(axis cs:8,0.289457576831375)
--(axis cs:10,0.249440898939529)
--(axis cs:12,0.225022675646294)
--(axis cs:14,0.205669183189061)
--(axis cs:16,0.201189189569105)
--(axis cs:18,0.200992182680649)
--(axis cs:20,0.201008130820037)
--(axis cs:20,0.220304615494284)
--(axis cs:20,0.220304615494284)
--(axis cs:18,0.220309878305269)
--(axis cs:16,0.221284576687062)
--(axis cs:14,0.225482651993709)
--(axis cs:12,0.245020132817475)
--(axis cs:10,0.27404253475677)
--(axis cs:8,0.323155453563942)
--(axis cs:6,0.436131302965591)
--(axis cs:4,0.504252658179787)
--(axis cs:2,0.637966936875157)
--cycle;

\path [fill=color2, fill opacity=0.05]
(axis cs:2,0.66822586139265)
--(axis cs:2,0.652253350908496)
--(axis cs:4,0.499891041910509)
--(axis cs:6,0.410888480986968)
--(axis cs:8,0.280976010451325)
--(axis cs:10,0.249579947432809)
--(axis cs:12,0.22450946312531)
--(axis cs:14,0.205705375222141)
--(axis cs:16,0.201219276909635)
--(axis cs:18,0.201011003771559)
--(axis cs:20,0.201121017625379)
--(axis cs:20,0.220249547568473)
--(axis cs:20,0.220249547568473)
--(axis cs:18,0.220306514915281)
--(axis cs:16,0.221293439323048)
--(axis cs:14,0.225479062834951)
--(axis cs:12,0.24463145747023)
--(axis cs:10,0.273550469103314)
--(axis cs:8,0.310724828552114)
--(axis cs:6,0.445056399014647)
--(axis cs:4,0.540278255220045)
--(axis cs:2,0.66822586139265)
--cycle;

\path [fill=color3, fill opacity=0.05]
(axis cs:2,0.676004266006413)
--(axis cs:2,0.667900535880923)
--(axis cs:4,0.511452759697139)
--(axis cs:6,0.418752151152457)
--(axis cs:8,0.285754653068405)
--(axis cs:10,0.250346432765294)
--(axis cs:12,0.223200668840241)
--(axis cs:14,0.205762160693734)
--(axis cs:16,0.201390202997722)
--(axis cs:18,0.201103449121809)
--(axis cs:20,0.201157933604656)
--(axis cs:20,0.220253865277293)
--(axis cs:20,0.220253865277293)
--(axis cs:18,0.220343382824096)
--(axis cs:16,0.221155761218685)
--(axis cs:14,0.225077718152446)
--(axis cs:12,0.243840237130204)
--(axis cs:10,0.274695634260741)
--(axis cs:8,0.313785628823351)
--(axis cs:6,0.449928596839354)
--(axis cs:4,0.547458761979679)
--(axis cs:2,0.676004266006413)
--cycle;

\path [fill=color4, fill opacity=0.05]
(axis cs:2,0.679959801360147)
--(axis cs:2,0.673344968037269)
--(axis cs:4,0.541310524045081)
--(axis cs:6,0.432019325767531)
--(axis cs:8,0.296875065892084)
--(axis cs:10,0.254665806737024)
--(axis cs:12,0.223385844676223)
--(axis cs:14,0.205837404794793)
--(axis cs:16,0.201544160827097)
--(axis cs:18,0.201209443338233)
--(axis cs:20,0.201402081846979)
--(axis cs:20,0.220739206589456)
--(axis cs:20,0.220739206589456)
--(axis cs:18,0.220375068991551)
--(axis cs:16,0.22109218290684)
--(axis cs:14,0.22489040952923)
--(axis cs:12,0.244777875658184)
--(axis cs:10,0.278974199510482)
--(axis cs:8,0.324865447719525)
--(axis cs:6,0.463074964084019)
--(axis cs:4,0.574268572687894)
--(axis cs:2,0.679959801360147)
--cycle;

\path [fill=color5, fill opacity=0.05]
(axis cs:2,0.680013053360247)
--(axis cs:2,0.67341971994272)
--(axis cs:4,0.541801580160565)
--(axis cs:6,0.432249259475154)
--(axis cs:8,0.2971370052027)
--(axis cs:10,0.254792104073547)
--(axis cs:12,0.223407445521715)
--(axis cs:14,0.205840270865644)
--(axis cs:16,0.201547316434022)
--(axis cs:18,0.201222786944813)
--(axis cs:20,0.201405709468905)
--(axis cs:20,0.220739491702744)
--(axis cs:20,0.220739491702744)
--(axis cs:18,0.220393123920801)
--(axis cs:16,0.22109138783213)
--(axis cs:14,0.224889376509679)
--(axis cs:12,0.244812478775808)
--(axis cs:10,0.27909492740389)
--(axis cs:8,0.325135547403848)
--(axis cs:6,0.46330034780847)
--(axis cs:4,0.574710992521842)
--(axis cs:2,0.680013053360247)
--cycle;

\addplot [thick, color0]
table {%
2 0.665088479939596
4 0.539938187925457
6 0.448619270707037
8 0.306435960423653
10 0.2683136067017
12 0.236791795067915
14 0.217495807227158
16 0.21302506554299
18 0.212153823659805
20 0.212178267145391
};
\addplot [thick, color1]
table {%
2 0.628248445583525
4 0.483398706011149
6 0.41890647901913
8 0.306306515197659
10 0.26174171684815
12 0.235021404231885
14 0.215575917591385
16 0.211236883128083
18 0.210651030492959
20 0.210656373157161
};
\addplot [thick, color2]
table {%
2 0.660239606150573
4 0.520084648565277
6 0.427972440000807
8 0.29585041950172
10 0.261565208268062
12 0.23457046029777
14 0.215592219028546
16 0.211256358116342
18 0.21065875934342
20 0.210685282596926
};
\addplot [thick, color3]
table {%
2 0.671952400943668
4 0.529455760838409
6 0.434340373995905
8 0.299770140945878
10 0.262521033513018
12 0.233520452985223
14 0.21541993942309
16 0.211272982108204
18 0.210723415972953
20 0.210705899440975
};
\addplot [thick, color4]
table {%
2 0.676652384698708
4 0.557789548366487
6 0.447547144925775
8 0.310870256805805
10 0.266820003123753
12 0.234081860167203
14 0.215363907162012
16 0.211318171866968
18 0.210792256164892
20 0.211070644218217
};
\addplot [thick, color5]
table {%
2 0.676716386651484
4 0.558256286341204
6 0.447774803641812
8 0.311136276303274
10 0.266943515738718
12 0.234109962148761
14 0.215364823687661
16 0.211319352133076
18 0.210807955432807
20 0.211072600585824
};
\end{axis}

\end{tikzpicture}
	    \vspace{-0.2cm}
	    \caption{NLPD}
    \end{subfigure}
	\hfill
	\begin{subfigure}[b]{.32\textwidth}
	    \centering
\begin{tikzpicture}

\definecolor{color0}{rgb}{0.12156862745098,0.466666666666667,0.705882352941177}
\definecolor{color1}{rgb}{1,0.498039215686275,0.0549019607843137}
\definecolor{color2}{rgb}{0.172549019607843,0.627450980392157,0.172549019607843}
\definecolor{color3}{rgb}{0.83921568627451,0.152941176470588,0.156862745098039}
\definecolor{color4}{rgb}{0.580392156862745,0.403921568627451,0.741176470588235}
\definecolor{color5}{rgb}{0.549019607843137,0.337254901960784,0.294117647058824}

\begin{axis}[
height=\figureheight,
tick pos=left,
width=\figurewidth,
x grid style={white!69.0196078431373!black},
xlabel={Number of inducing inputs, \(\displaystyle M\)},
xmin=2, xmax=16,
xtick style={color=black},
y grid style={white!69.0196078431373!black},
ymin=0.0666993342966855, ymax=0.484507498655418,
ytick style={color=black}
]
\path [fill=color0, fill opacity=0.05]
(axis cs:2,0.425660377358491)
--(axis cs:2,0.393962264150943)
--(axis cs:4,0.225864444884833)
--(axis cs:6,0.218253981497034)
--(axis cs:8,0.112316724613208)
--(axis cs:10,0.104478135624335)
--(axis cs:12,0.0909433962264151)
--(axis cs:14,0.0857377127447122)
--(axis cs:16,0.0874921423312971)
--(axis cs:18,0.0892636516713146)
--(axis cs:20,0.0882032602700912)
--(axis cs:20,0.102740135956324)
--(axis cs:20,0.102740135956324)
--(axis cs:18,0.102811820026799)
--(axis cs:16,0.103073895404552)
--(axis cs:14,0.104073608010005)
--(axis cs:12,0.108301886792453)
--(axis cs:10,0.121559600224722)
--(axis cs:8,0.129192709349056)
--(axis cs:6,0.263632810955796)
--(axis cs:4,0.260550649454789)
--(axis cs:2,0.425660377358491)
--cycle;

\path [fill=color1, fill opacity=0.05]
(axis cs:2,0.407828011482195)
--(axis cs:2,0.371039913046107)
--(axis cs:4,0.291194033366194)
--(axis cs:6,0.264398689970613)
--(axis cs:8,0.121667319731513)
--(axis cs:10,0.105729273875747)
--(axis cs:12,0.093877830786356)
--(axis cs:14,0.0866479134771323)
--(axis cs:16,0.0880330886123743)
--(axis cs:18,0.0891666841758982)
--(axis cs:20,0.0880171210941399)
--(axis cs:20,0.103303633622841)
--(axis cs:20,0.103303633622841)
--(axis cs:18,0.103286146012781)
--(axis cs:16,0.103665024595173)
--(axis cs:14,0.103540765768151)
--(axis cs:12,0.109141037138172)
--(axis cs:10,0.12219525442614)
--(axis cs:8,0.141351548193015)
--(axis cs:6,0.297488102482217)
--(axis cs:4,0.325032381728145)
--(axis cs:2,0.407828011482195)
--cycle;

\path [fill=color2, fill opacity=0.05]
(axis cs:2,0.426168058026428)
--(axis cs:2,0.387794206124515)
--(axis cs:4,0.24887225248932)
--(axis cs:6,0.222395157348744)
--(axis cs:8,0.112871437037885)
--(axis cs:10,0.104192058005842)
--(axis cs:12,0.0931525196897877)
--(axis cs:14,0.0863399551224029)
--(axis cs:16,0.0880330886123743)
--(axis cs:18,0.0891666841758982)
--(axis cs:20,0.0882536654994753)
--(axis cs:20,0.103067089217506)
--(axis cs:20,0.103067089217506)
--(axis cs:18,0.103286146012781)
--(axis cs:16,0.103665024595173)
--(axis cs:14,0.103471365632314)
--(axis cs:12,0.108734272763042)
--(axis cs:10,0.121468319352649)
--(axis cs:8,0.133543657301738)
--(axis cs:6,0.252699182273897)
--(axis cs:4,0.289995672038982)
--(axis cs:2,0.426168058026428)
--cycle;

\path [fill=color3, fill opacity=0.05]
(axis cs:2,0.465349239911863)
--(axis cs:2,0.430877175182477)
--(axis cs:4,0.29032232866098)
--(axis cs:6,0.279479531999685)
--(axis cs:8,0.115689906631015)
--(axis cs:10,0.101336203609248)
--(axis cs:12,0.0909033421629337)
--(axis cs:14,0.0856906144948097)
--(axis cs:16,0.0877289492344592)
--(axis cs:18,0.0891834838125521)
--(axis cs:20,0.0874012835569036)
--(axis cs:20,0.103164754178945)
--(axis cs:20,0.103164754178945)
--(axis cs:18,0.103646704866693)
--(axis cs:16,0.103214446991956)
--(axis cs:14,0.104498064750473)
--(axis cs:12,0.1087192993465)
--(axis cs:10,0.122060022805846)
--(axis cs:8,0.133744055633136)
--(axis cs:6,0.31146386422673)
--(axis cs:4,0.328545595867322)
--(axis cs:2,0.465349239911863)
--cycle;

\path [fill=color4, fill opacity=0.05]
(axis cs:2,0.465516218457294)
--(axis cs:2,0.431087555127612)
--(axis cs:4,0.283353055129638)
--(axis cs:6,0.279022595300745)
--(axis cs:8,0.117952872769348)
--(axis cs:10,0.102250184275777)
--(axis cs:12,0.0916611881236451)
--(axis cs:14,0.0868682288546134)
--(axis cs:16,0.0867326340354112)
--(axis cs:18,0.0888703438907102)
--(axis cs:20,0.0890856888303566)
--(axis cs:20,0.103744499848889)
--(axis cs:20,0.103744499848889)
--(axis cs:18,0.103582486297969)
--(axis cs:16,0.103456045209872)
--(axis cs:14,0.105584601334066)
--(axis cs:12,0.110980321310317)
--(axis cs:10,0.122655476101582)
--(axis cs:8,0.136764108362728)
--(axis cs:6,0.31569438583133)
--(axis cs:4,0.332873359964702)
--(axis cs:2,0.465516218457294)
--cycle;

\path [fill=color5, fill opacity=0.05]
(axis cs:2,0.465516218457294)
--(axis cs:2,0.431087555127612)
--(axis cs:4,0.283004895516394)
--(axis cs:6,0.279022595300745)
--(axis cs:8,0.117952872769348)
--(axis cs:10,0.102250184275777)
--(axis cs:12,0.0916611881236451)
--(axis cs:14,0.086549214369298)
--(axis cs:16,0.0867326340354112)
--(axis cs:18,0.0888703438907102)
--(axis cs:20,0.0890856888303566)
--(axis cs:20,0.103744499848889)
--(axis cs:20,0.103744499848889)
--(axis cs:18,0.103582486297969)
--(axis cs:16,0.103456045209872)
--(axis cs:14,0.105526257328815)
--(axis cs:12,0.110980321310317)
--(axis cs:10,0.122655476101582)
--(axis cs:8,0.136764108362728)
--(axis cs:6,0.31569438583133)
--(axis cs:4,0.333976236559078)
--(axis cs:2,0.465516218457294)
--cycle;

\addplot [thick, color0]
table {%
2 0.409811320754717
4 0.243207547169811
6 0.240943396226415
8 0.120754716981132
10 0.113018867924528
12 0.099622641509434
14 0.0949056603773585
16 0.0952830188679245
18 0.0960377358490566
20 0.0954716981132075
};
\addplot [thick, color1]
table {%
2 0.389433962264151
4 0.30811320754717
6 0.280943396226415
8 0.131509433962264
10 0.113962264150943
12 0.101509433962264
14 0.0950943396226415
16 0.0958490566037736
18 0.0962264150943396
20 0.0956603773584906
};
\addplot [thick, color2]
table {%
2 0.406981132075472
4 0.269433962264151
6 0.237547169811321
8 0.123207547169811
10 0.112830188679245
12 0.100943396226415
14 0.0949056603773585
16 0.0958490566037736
18 0.0962264150943396
20 0.0956603773584906
};
\addplot [thick, color3]
table {%
2 0.44811320754717
4 0.309433962264151
6 0.295471698113208
8 0.124716981132075
10 0.111698113207547
12 0.099811320754717
14 0.0950943396226415
16 0.0954716981132075
18 0.0964150943396226
20 0.0952830188679245
};
\addplot [thick, color4]
table {%
2 0.448301886792453
4 0.30811320754717
6 0.297358490566038
8 0.127358490566038
10 0.112452830188679
12 0.101320754716981
14 0.0962264150943396
16 0.0950943396226415
18 0.0962264150943396
20 0.0964150943396226
};
\addplot [thick, color5]
table {%
2 0.448301886792453
4 0.308490566037736
6 0.297358490566038
8 0.127358490566038
10 0.112452830188679
12 0.101320754716981
14 0.0960377358490566
16 0.0950943396226415
18 0.0962264150943396
20 0.0964150943396226
};
\end{axis}

\end{tikzpicture}
	    \vspace{-0.2cm}
	    \caption{Classification Error}
	\end{subfigure}
	\vspace{-0.25cm}
  \caption{The Banana 2D classification task with varying number of inducing inputs (mean and standard deviation of 10-fold cross-validation). All methods improve monotonically as $M$ increases. \sspep($\alpha=1$) and \sspl slightly outperform the other methods when $M$ is small. \sscvi and \sspep($\alpha=0.01$) perform identically.
  }
  \label{fig:vary_M}
  \vspace{-0.25cm}
\end{figure*}
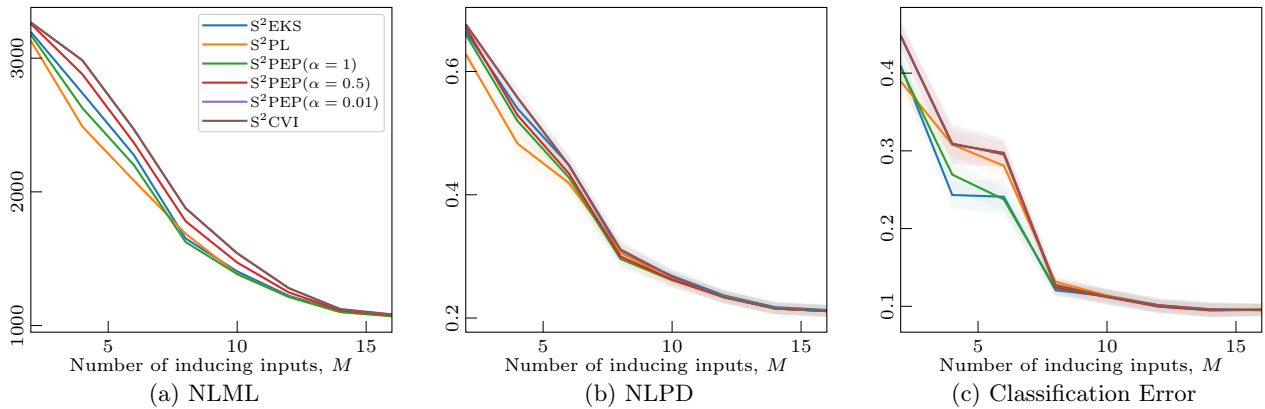

\subsection{Doubly Sparse Posterior Linearisation (\sspl) and Nonlinear Kalman Smoothers}

Site-based inference is in fact more general than just PEP and CVI. \citet{wilkinson2020state} showed that classical nonlinear Kalman smoothers, such as the Extended, Unscented and Gauss-Hermite smoothers, can also be formulated as site-based algorithms. These algorithms are based on various forms of linearisation of the likelihood model, and their approach is generalised and improved upon in a method called \emph{posterior linearisation} \citep[PL,][]{garcia2016iterated}. We derived a sparse extension to the posterior linearisation algorithms presented in \citet{wilkinson2020state}, including a sparse version of the extended Kalman smoother (\sseks). Details of the derivations are given in \cref{app:slr}.

\subsection{Algorithmic Details}
\paragraph{Approximate marginal likelihood} For all our algorithms, the marginal likelihood can be written as: ${p(\vy) = p(y_1)\,p(y_2 \mid y_1)\,p(y_3 \mid \vy_{1:2})\prod_{n=4}^N p(y_n \mid \vy_{1:n-1})}$, and each term can be approximated during a forward filter pass through the data by noticing that,
\begin{align} \label{eq:marg-lik}
 &p(y_n \mid \vy_{1:n-1}) = \nonumber\\
&\mint p(y_n \mid f_n = \vH\vs(x_n)) \, \tilde{p}(\vs(x_n) \mid \vy_{1:n-1}) \dd\vs(x_n), 
\end{align} 
where $\tilde{p}(\vs(x_n) \mid \vy_{1:n-1})$
is an approximate forward filter prediction calculated by replacing the likelihood term by the sites when filtering over the inducing states.

Alternative approximations to the marginal likelihood also exist. The PEP energy is obtained by marginalizing the approximate joint $\cZ_{pep} = \mint q(\vs(\cdot)) \dd \vs(\cdot)$, as described in \citet{bui2017unifying} and in \cref{app:sec:pep_energy}. For CVI, the ELBO is typically used in place of the marginal likelihood (see \cref{app:sec:elbo}), and as with the PEP energy all its terms can be computed in $\bigO(Md^3)$.

\paragraph{Parallelizing the updates} 
The sites may be updated one at a time as described, or they can be updated simultaneously as is done in parallel EP \citep{li2015stochastic}. This particular setting is the closest to \ssvgp in terms of both posterior approximation structure, storage and computational complexity. The ability to perform site updates in batches also facilitates stochastic optimisation, leading to overall computational complexity of $\bigO((M+N_*)d^3)$ for batch size $N_*$.

\subsection{Spatio-temporal Gaussian Processes}

As with standard filtering approaches to inference, our doubly sparse approach is compatible with spatio-temporal GP models \citep{Sarkka+Solin+Hartikainen:2013, tebbutt2021combining}, allowing for analysis of data sets with input dimension greater than one.
Here, we construct a sparse spatio-temporal GP, $f(x,\vr)$, with inducing inputs in space, $\vz_r$, indexing a finite set of coupled inducing temporal GPs, $\vs(x)$, and we also impose these temporal GPs to be sparse with inducing states $\vu$ indexed at temporal inputs $\vz_x$.
\cref{fig:banana} shows a demonstration of this approach on two-dimensional classification tasks. 

We focus on the case of separable stationary spatio-temporal kernels where ${\kappa(x,\vr,x',\vr')} = {\kappa_x(x-x')}\,{\kappa_r(\vr- \vr')}$ and where $\kappa_x$ is Markovian with state dimension $d$. A spatio-temporal GP with such a kernel has an equivalent representation as a stochastic partial differential equation (SPDE),
\begin{gather}
\begin{aligned}    \frac{\partial\vcs(x,\cdot)}{\partial x} &= \mathcal{A}_r \vcs(x, \cdot) + \mathcal{L}_r \vw(x, \cdot),\\
    f(x,\vr) &= \vH_x \vcs(x, \vr).
\end{aligned}
\end{gather}
This infinite-dimensional SPDE marginalized to a finite set of $M_z$ spatial locations $\vz_r$ is a finite-dimensional SDE with state dimension $M_z d$. Noting $\vs(x)=\vcs(x, \vz_r)$ we have, 
\begin{align}
 \label{eq:spatio-temp-sde}
    \frac{\mathrm{d}\vs(x)}{\mathrm{d} x} = \vF \vs(x) + \vL \veps(x).
\end{align}
The parameters of the SDE in \cref{eq:spatio-temp-sde} are given in \citet{Sarkka+Solin+Hartikainen:2013}. Intuitively, state $\vs$ splits into $M_z$ correlated temporal processes $\vs = [\vs_1, \dots \vs_{M_z}]$, whose marginal projection ${f_i(\cdot) = \vH_x \vs_i(\cdot)}$ verify
\begin{gather}
\begin{aligned} 
\mathbb{C}\mathrm{ov}[f_i(x), f_i(x')] &= \kappa_x(x-x')\kappa_r(0),\\ 
\mathbb{C}\mathrm{ov}[f_i(x), f_j(x)] &= \kappa_x(0) \kappa_r(\vz^i_r- \vz^j_r).
\end{aligned}
\end{gather}
We can further marginalise this SDE to its values ${\vu=\vs(\vz_x)}$ at $M_x$ temporal inputs $\vz_x$, leading to the discrete state-space model, 
\begin{gather}
\begin{aligned} 
\vu_0 &\sim \NN(\bm{0}, \vK^r_{\vz_r,\vz_r} \otimes \vP^x_{0}), \\
    \vu_{m+1} &= \vA_{m,{m+1}} \, \vu_{m} + \vq_m,
\end{aligned}
\end{gather}
where $\vP^x_{0}$ is the stationary covariance of $\vs$ in \cref{eq:spatio-temp-sde}. In our sparse algorithms for spatio-temporal models, we use $\vu$ as inducing variables and we need the conditional $p(f_n\mid\vu)$ to make predictions about the process.

For a single data point, $(x_n, \vr_n, y_n)$, and denoting ${f_n=f(x_n, \vr_n)}$, there is a conditional independence property specific to separable kernels: ${p(f_n\mid\vs(\cdot)) = p(f_n\mid\vs(x_n))}$  \citep[see][]{tebbutt2021combining}.
The conditional $p(f_n\mid\vu)$ is then obtained by marginalizing $\vs(x_n)$ in the joint ${p(f_n, \vs(x_n) \mid \vu) = p(f_n\mid \vs(x_n))p(\vs(x_n) \mid \vu)}$, given by
\begin{gather}
\begin{aligned}     
p(\vs(x_n) \mid \vu) &= \NN(\vs(x_n) \mid \vR_n \, \vv_{m(n)}, \vT_n), \\ 
    p(f_n \mid \vs(x_n)) &= \NN(f_n \mid \vB(\vr_n) \, \vs(x_n), \vC(\vr_n)), %
\end{aligned}
\end{gather}

where $\vR_n$ and $\vT_n$ are defined as in \cref{app:conditionals} and 
\begin{gather}
\begin{aligned}     \vB(\vr_n)&=\left[\vK_{\vr_n,\vz_r} \vK_{\vz_r,\vz_r}^{-1} \right] \otimes \vH,\\
    \vC(\vr_n) &= \kappa_x(0) \left( \vK_{\vr_n,\vr_n} - \vK_{\vr_n,\vz_r} \vK_{\vz_r,\vz_r}^{-1} \vK_{\vz_r, \vr_n} \right) .
\end{aligned}
\end{gather}

There are several approximate inference approaches based on the SPDE formulation. {\em (i)}~In \citet{simpson2012think}, the SPDE is approximated via a local basis expansion where the associated weights are distributed as a Gaussian Markov random field. Its sparse precision matrix leads to efficient computations. However, the generative model (prior) is approximated, which is not the case in our approach. {\em (ii)}~Global approximations based on the SPDE formulation \citep[\eg,][]{solin2020hilbert} also approximate the prior based on the truncation of an exact infinite expansion of the kernel. Our approach singles out a time dimension which turns the SPDE into a SDE with an infinite-dimensional state \citep{Sarkka+Solin+Hartikainen:2013}. Using a further sparse approximation to this infinite dimensional state leads to inference in an SDE with finite dimension.%

\begin{table*}[t]
	\caption{Normalised negative log predictive density (NLPD) results using 10-fold cross-validation. Mean and standard deviation shown (smaller is better). The banded matrix operations currently used for filtering in \ssvgp are unstable for large datasets, while SVGP does not scale to more than $\sim$\,1000 inducing points.} 
	\vspace*{-1em}
	\label{tbl:results}
	\renewcommand{\arraystretch}{.75}
	\tikzstyle{every picture}+=[remember picture]
	\tikzstyle{na} = [baseline=-.5ex]
	\begin{center}
		{\fontsize{7.15pt}{10.5pt} 
			\selectfont
			\setlength{\tabcolsep}{0pt}
			\newlength{\tblw}
			\setlength{\tblw}{0.12\textwidth}
			\begin{sc}
				\begin{tabularx}{\textwidth}{p{1.35\tblw} C{\tblw} C{\tblw} C{\tblw} C{\tblw} C{\tblw} C{\tblw} C{\tblw}}
					\toprule
					& Motorcycle & Coal & Banana & Binary & Audio & Airline & Electricity  \\
					\midrule
					\# data points, $N$ & 133 & 333 & 5300 & 10k & 22k & 36k & 262k \\
					\# inducing inputs, $M$ & 30 & 15 & 15\,$\times$\,15 & 1k & 3k & 4k & 50k \\
					Input dimension & 1 & 1 & 2 & 1 & 1 & 1 & 1 \\
					Likelihood & Heteroscedastic & Poisson & Bernoulli & Bernoulli & Product & Poisson & Gaussian \\
					\midrule 
					\sseks & 0.870$\pm$0.16 & 0.924$\pm$0.11 & 0.212$\pm$0.01 & 0.205$\pm$0.02 & 0.218$\pm$0.00 & 0.128$\pm$0.04 & $-$0.085$\pm$0.02 \\
					\sspl & 0.892$\pm$0.15 & 0.925$\pm$0.11 & 0.211$\pm$0.01 & 0.189$\pm$0.02 & 0.213$\pm$0.11 & 0.128$\pm$0.04 & $-$0.085$\pm$0.02 \\
					\sspep ($\alpha=1$) & 0.456$\pm$0.37 & 0.924$\pm$0.11 & 0.211$\pm$0.01 & 0.189$\pm$0.02 & \textbf{$-$1.326$\pm$0.01} & 0.128$\pm$0.04 & $-$0.085$\pm$0.02 \\
					\sspep ($\alpha=0.5$) & \textbf{0.420$\pm$0.35} & 0.924$\pm$0.11 & 0.211$\pm$0.01 & 0.189$\pm$0.02 & $-$0.624$\pm$0.07 & 0.128$\pm$0.04 & $-$0.074$\pm$0.02 \\
					\sspep ($\alpha=0.01$) & 0.428$\pm$0.33 & 0.924$\pm$0.11 & 0.211$\pm$0.01 & \textbf{0.188$\pm$0.02} & 0.624$\pm$0.04 & 0.128$\pm$0.04 & \textbf{$-$0.153$\pm$0.01} \\
					\sscvi & 0.428$\pm$0.33 & 0.924$\pm$0.11 & 0.211$\pm$0.01 & \textbf{0.188$\pm$0.02} & 0.681$\pm$0.03 & 0.128$\pm$0.04 & $-$0.152$\pm$0.01 \\
					\midrule
					\ssvgp & 0.434$\pm$0.31 \ & 0.937$\pm$0.10 & 0.215$\pm$0.01 & 0.236$\pm$0.01 & $\times$ & $\times$ & $\times$ \\
					\addlinespace[0.15em]
					Sparse VI (SVGP) & 0.440$\pm$0.30 & 0.954$\pm$0.12 & 0.226$\pm$0.01  & 0.207$\pm$0.02 & $\times$ & $\times$ & $\times$ \\
					\midrule
					\addlinespace[0.15em]
					Full EKS & 0.871$\pm$0.16 & 0.924$\pm$0.11 & 0.212$\pm$0.01 & 0.205$\pm$0.02 & 0.412$\pm$0.01 & 0.128$\pm$0.04 & $-$0.420$\pm$0.01 \\
					Full PL & 0.893$\pm$0.15 & 0.924$\pm$0.12 & 0.211$\pm$0.01 &0.189$\pm$0.02 & $-$0.514$\pm$0.18& 0.128$\pm$0.04 & $-$0.420$\pm$0.01 \\
					Full PEP ($\alpha=1$) & 0.429$\pm$0.31 & 0.924$\pm$0.12 & 0.211$\pm$0.01 & 0.189$\pm$0.02 & $-$1.441$\pm$0.02 & 0.128$\pm$0.04 & $-$0.420$\pm$0.01 \\
					Full PEP ($\alpha=0.5$) & 0.422$\pm$0.31 & 0.924$\pm$0.12 & 0.211$\pm$0.01 & 0.189$\pm$0.02 & $-$0.902$\pm$0.05 & 0.128$\pm$0.04 & $-$0.420$\pm$0.01 \\
					Full PEP ($\alpha=0.01$) & 0.416$\pm$0.32 & 0.924$\pm$0.12 & 0.211$\pm$0.01 & 0.188$\pm$0.02 & 0.169$\pm$0.08 & 0.128$\pm$0.04 & $-$0.420$\pm$0.01 \\
					Full CVI & 0.415$\pm$0.32 & 0.924$\pm$0.12 & 0.211$\pm$0.01 &0.188$\pm$0.02 & 0.671$\pm$0.03& 0.128$\pm$0.04 & $-$0.420$\pm$0.01 \\
					\bottomrule
				\end{tabularx}
			\end{sc}						
		}	
	\end{center}
	\vspace*{-1em}
\end{table*}

\section{EMPIRICAL ANALYSIS}
\label{sec:results}

In \cref{fig:vary_M} we analyse the effect of increasing the number of inducing inputs, $M$, in a two-dimensional classification task. We observe that the training marginal likelihood (NLML), the test predictive density (NLPD), and the test classification error improve monotonically as $M$ increases, as expected. When $M$ is very small, the methods which use the EP energy for training (\sspep($\alpha=1$), \sspl, \sseks) perform well. We provide a similar analysis of the more complicated Audio task in \cref{app:experiment-details}, in which \sspep again requires few inducing inputs to obtain good results. %

\cref{tbl:results} analyses the practical performance of our site-based algorithms relative to the \ssvgp approach, and compared to their full (non-sparse) site-based equivalents. We use six of the benchmark tasks presented in \citet{wilkinson2020state}, and add the additional Electricity task in order to show that the methods are applicable to extremely large time series. See \cref{app:experiment-details} for further details on the data sets and full models used. Four of the data sets are sufficiently small that we are also able to compare against standard sparse VI \citep[SVGP, using GPflow,][]{GPflow:2017}.

We measure the negative log predictive density (NLPD), using 10-fold cross-validation, with each method run for 500 training iterations. Each iteration consists of an update to the sites, followed by a gradient step to update the hyperparameters using Adam. The ELBO is used as the training objective for the variational methods, and the power EP energy for PEP (see \cref{app:sec:pep_energy}). The PL and EKS methods also use the PEP energy ($\alpha=1$, see \cref{app:slr} for discussion). In the non-conjugate tasks, we use Gauss--Hermite integration with $20^q$ cubature points, where $q$ is the dimensionality of the integral being approximated. However, in the Audio task this approach is not practical, and so we use the fifth-order Unscented transform \citep{mcnamee1967construction}. %

\paragraph{Results} We observe the performance to be highly model and task dependent. For many tasks, the methods all perform similarly. However, it is worth noting that \sspep performs well on the difficult Audio task which has the most complicated likelihood model. However, \sscvi sometimes outperforms \sspep when the likelihood is simpler (\eg, the Binary classification task, and the Electricity regression task). \cref{fig:banana} illustrates the performance of the doubly sparse methods in comparison to SVGP on the 2D classification task. In the heteroscedastic noise task, \sspep is the best performing sparse method. As expected, \sspep($\alpha=0.01$) and \sscvi give similar results, except in the Audio task, where numerical integration error resulting from the three-dimensional cubature used for the updates causes the results to differ.

\section{CONCLUSIONS}
\label{sec:discussion}

We have derived site-based inference methods for sparse Markovian GPs. In doing so, we have shown the generality of the sparse Markovian approach, and provided a suite of algorithms applicable to large temporal data. We also proposed a principled approach to site tying motivated by the specific structure of the prior on inducing states, resulting in methods with very efficient computational and memory scaling. %

The site-based approach makes it possible to apply PEP (as well as the classical Kalman smoothers) in the doubly sparse framework, and this method outperforms existing approaches on some difficult non-conjugate tasks. Our new algorithms inherit many of the desirable properties of their full counterparts, including the ability to handle spatio-temporal models, resulting in a novel sparse approach in which inducing points in time and space are fully decoupled.  %

\section*{Acknowledgements}
\noindent
We wish to thank the NVIDIA AI Technology Center (NVAITC) Finland, specifically Niki Loppi, who provided invaluable help in developing efficient code for this project.
We also acknowledge the computational resources provided by the Aalto Science-IT project. We acknowledge funding from the Academy of Finland (grant number 324345).

\clearpage

\phantomsection%
\addcontentsline{toc}{section}{References}
\begingroup
\bibliographystyle{abbrvnat}
\bibliography{main}
\endgroup

\def\toptitlebar{\hrule height4pt \vskip .25in \vskip -\parskip} 
\def\bottomtitlebar{\vskip .29in \vskip -\parskip \hrule height1pt \vskip .09in} 

\newcommand{\nipstitle}[1]{{%
    \phantomsection\hsize\textwidth\linewidth\hsize%
    \vskip 0.1in%
    \toptitlebar%
    \begin{minipage}{\textwidth}%
        \centering{\Large\bf #1\par}%
    \end{minipage}%
    \bottomtitlebar%
    \addcontentsline{toc}{section}{#1}%
}}

\clearpage

\onecolumn

\appendix

\clearpage

\setcounter{section}{0}

\nipstitle{
    {Supplementary Material:}\\ 
    Sparse Algorithms for Markovian Gaussian Processes}

\pagestyle{empty}

\section{Statistical Properties of Linear SDEs} \label{app:sde}

\subsection{Marginals} \label{app:marginals}

A linear time invariant (LTI) stochastic differential equation (SDE) can be expressed as follows:
\begin{align}
    \dot{\vs}(x) &= \vF \vs(x) + \vL \veps(x)\,, \qquad
    f(x) = \vH \vs(x)\,,
\end{align}
where $\veps(x)$ is a white noise process, $\vF$ is the feedback matrix, $\vL$ is the noise effect matrix, and $\vH$ is the measurement matrix. 

The marginal distribution of the solution to this LTI-SDE evaluated at any ordered set $\vx = [x_1,\dots, x_N]\in \RR^N$ follows a discrete-time linear system:
\begin{gather}
\begin{aligned}
    \vs(x_{n+1}) &= \vA_{n,{n+1}} \vs(x_n) + \vq_n, \quad\quad\quad & \vq_n &\sim \NN(\bm{0}, \vQ_{n,{n+1}}), \\
    \vs(x_0) &\sim \NN(\bm{0}, \vP_0), & \hspace*{-1em} f_n &= \vH \vs(x_n),
\end{aligned}
\end{gather}
where the state transition matrices, $\vA_{n,{n+1}} \in \RR^{d\times d}$, noise covariance matrices, $\vQ_{n,{n+1}} \in \RR^{d\times d}$, and stationary state covariance matrix $\vP_0 \in \RR^{d\times d}$ can be computed analytically. Denoting the matrix exponential as $\vPhi$ and with step size $\Delta_n = x_{n+1} - x_{n}$, we have
\begin{gather}
\begin{aligned}
    \hspace*{-0.6em}\vA_{
n,{n+1}} &= \vPhi( \vF \Delta_n)\,, \qquad \\
    \hspace*{-0.6em}\vQ_{n,{n+1}} &=  \mint_{0}^{\Delta_n} \vPhi(\Delta_n-\tau)\vL \vQ_c \vL^{\top} \vPhi(\Delta_n-\tau)^{\top} \dd\tau\,. 
\end{aligned}
\end{gather}

\subsection{Conditionals} \label{app:conditionals}

This section is adapted from Appendix~A.1 of \cite{adam2020doubly}. We consider a stationary Markovian GP with state dimension $d$ and denote by $(\vu_{\om}, \vs, \vu_{\op})$ its evaluation on the triplet $(z_\om, x, z_\op)$.
We here detail the derivation of $p(\vs \mid \vv=[\vu_{\om}, \vu_{\op}])$.
\subsubsection*{Derivation from the joint precision}
\begin{align}
p(\vs \mid \vu_{\om},  \vu_{\op}) &\propto p(\vs \mid \vu_{\om})p(\vu_{\op}\mid\vs) \nonumber\\
&\propto \NN (\vs; \vA_{\om,x} \vu_{\om}, \vQ_{\om,x}) \NN(\vu_{\op};\vA_{x,\op} \vs, \vQ_{x,\op}) \nonumber\\
&\propto  \exp -\frac{1}{2}
\left[  \|\vs - \vA_{\om,x} \vu_{\om}\|^2_{\vQ_{\om,x}^{-1}}  + \|\vu_{\op} - \vA_{x,\op} \vs\|^2_{\vQ_{x,\op}^{-1}} \right] \nonumber\\
&\propto  \exp -\frac{1}{2}\big[ 
     \vs^\top \underset{\vT^{-1}}{\underbrace{(\vQ_{\om,x}^{-1} + (\vA_{x,\op})^\top \vQ_{x,\op}^{-1} \vA_{x,\op})}} \vs  - 2 \vs^\top 
     \underset{\vM = [\vM_1, \vM_2]}{\underbrace{\begin{bmatrix} \vQ_{\om,x}^{-1} \vA_{\om,x}, &  \vA^\top_{x,\op} \vQ_{x,\op}^{-1}  \end{bmatrix}}} \vv  \big] \nonumber\\
&\propto  \exp -\frac{1}{2}\left[ \vs^\top \vT^{-1} \vs - 2 \vs^\top  \vM \vv   \right]  = \NN(\vs; \vR \vv, \vT) 
\end{align}
with
\begin{align}
\vT &=  (\vQ_{\om,x}^{-1} + \vA^\top_{x,\op} \vQ_{x,\op}^{-1} \vA_{x,\op})^{-1} \; \text{(Woodbury identity)}\nonumber\\
&=  \vQ_{\om, x} -  \vQ_{\om,x} \vA^\top_{x,\op} ( \vQ_{x,\op} + \vA_{x,\op} \vQ_{\om,x}\vA_{x,\op}^\top)^{-1}\vA_{x,\op}  \vQ_{\om,x}\nonumber\\
&=  \vQ_{\om, x} -  \vQ_{\om,x} \vA^\top_{x,\op} \vQ_{\om,\op}^{-1}  \vA_{x,\op}  \vQ_{\om,x}
\end{align}
and
$    \vR = [\vR_1, \vR_2] = \vT \vM = [ \vT\vM_1,  \vT\vM_2] $ given by
\begin{align}
\vR_1 &= (\vQ_{\om, x} -  \vQ_{\om,x} \vA^\top_{x,\op} \vQ_{\om,\op}^{-1}  \vA_{x,\op}  \vQ_{\om,x}) \vQ_{\om,x}^{-1} \vA_{\om,x}  \nonumber\\
&= \vA_{\om,x}  -  \vQ_{\om,x} \vA^\top_{x,\op} \vQ_{\om,\op}^{-1}   \vA_{\om,\op},  \\
\vR_2 &= (\vQ_{\om, x} -  \vQ_{\om,x} \vA^\top_{x,\op} \vQ_{\om,\op}^{-1}  \vA_{x,\op}  \vQ_{\om,x}) \vA^\top_{x,\op} \vQ_{x,\op}^{-1}  \nonumber\\
&=\vQ_{\om, x} \vA^\top_{x,\op} \vQ_{x,\op}^{-1} -   \vQ_{\om, x} \vA^\top_{x,\op}\vQ_{\om,\op}^{-1} (\vQ_{\om,\op} - \vQ_{x,\op})  \vQ_{x,\op}^{-1} \; \text{(Woodbury identity)}\nonumber\\
&=  \vQ_{\om, x}\vA^\top_{x,\op}\vQ_{\om,\op}^{-1}.
\end{align}
The conditional function evaluation $f(x)  =\vH \vs$ is thus: 
\begin{align}
    p(f(x) \mid \vu_{\om}, \vu_{\op})  =\NN(f(x); \vH \vR \vv, \vH \vT \vH^\top)=  \NN(f(x); \vW \vv, \nu).
\end{align}

\section{Inference in Site-based Sparse Markovian GP Models} \label{app:inference}

The site based algorithms build an approximation to the posterior of the form:
\begin{align}
     q(\vs(\cdot)) \propto p(\vu)\,p(\vs(\cdot)\mid\vu)\mprod_m t_m(\vv_m).
\end{align}
The factors $t_m$ are called \emph{sites} and are parameterized as unnormalized Gaussian distributions in the natural parameterization: $t_m(\vv_m)= z_m \exp( \vv_m^\top \vT_{1,m} - \nicefrac{1}{2}\, \vv_m^\top \vT_{2,m}\vv_m) = \tNN(\vv_m; z_m, \vT_{1,m}, \vT_{2,m})$.
\subsection{Filtering and Smoothing}
\label{app:filtering}
It is possible to compute the posterior marginals over the individual inducing states $q(\vu_m)$ and pairwise consecutive inducing states $q(\vv_m=[\vu_m, \vu_{m+1}])$ by introducing the forward ($f$) and backward ($b$) filters: 
\begin{gather}
\begin{aligned}
    q^f(\vu_m) &\propto \mint p( \vu_{\leq m}) \mprod_{m'< m} t_{m'}(\vv_{m'}) \dd \vu_{<m} , \\ 
    q^b(\vu_{m}) &\propto \mint p(\vu_{\geq m}) \mprod_{m'\geq m} t_{m'}(\vv_{m'})  \dd \vu_{>m}.
\end{aligned}
\end{gather}
These can be evaluated using the following recursions: %
\begin{gather}
\begin{aligned}
    q^f(\vu_{m+1}) &= \mint p( \vu_{\leq {m+1}}) \mprod_{m'< {m+1}} t_{m'}(\vv_{m'}) \dd \vu_{<{m+1}}\\
    &= \mint p( \vu_{\leq m}) \mprod_{m'< m} t_{m'}(\vv_{m'})  \mint p( \vu_{m+1}\mid \vu_m) t_{m}(\vv_{m}) \dd \vu_{<{m+1}}\\
    &= \mint  \big[ \mint  p( \vu_{\leq m}) \mprod_{m'< m} t_{m'}(\vv_{m'}) \dd \vu_{<{m}} \big] p( \vu_{m+1}\mid \vu_m) t_{m}(\vv_{m}) \dd \vu_{m} \\
    &=\mint q^f(\vu_{m})\,p(\vu_{m+1}\mid\vu_{m})\,t_m(\vv_{m})\dd \vu_{m} , \\ 
    q^b(\vu_{m})
    &=
    \mint p(\vu_{\geq m}) \mprod_{m'\geq m} t_{m'}(\vv_{m'})  \dd \vu_{>m}.\\
    &=
    \mint \big[\mint p(\vu_{\geq m+1}) \mprod_{m'\geq m+1} t_{m'}(\vv_{m'}) \dd \vu_{>m+1} \big] p(\vu_{m}\mid \vu_{m+1})t_m(\vv_m) \dd \vu_{m+1}.\\
    &=\mint 
        q^b(\vu_{m+1})\,p( \vu_{m}\mid\vu_{m+1}) \, t_{m}(\vv_{m})  \dd \vu_{m+1}\\
        &=\mint 
        q^b(\vu_{m+1})\,p( \vu_{m+1}\mid\vu_{m})\, p(\vu_m) \,/\, p(\vu_{m+1}) \, t_{m}(\vv_{m})  \dd \vu_{m+1}. 
\end{aligned}
\end{gather}
The desired marginals are then obtained as the product of the forward and backward filtering distributions, divided by the prior: %
\begin{gather}
\begin{aligned}
    q^s(\vu_m) 
    &= \mint q(\vu) \dd \vu_{\neq m}\\
    &= \mint p(\vu) \prod_{m'} t_{m'}(\vv_{m'}) \dd \vu_{\neq m}\\
    &= \mint \big[ p(\vu_{\leq m}) \prod_{m'<m} t_{m'}(\vv_{m'})\big] \big[ p(\vu_{>m}\mid \vu_m) \prod_{m'\geq m} t_{m'}(\vv_{m'})\big] \dd \vu_{\neq m}\\
    &=  \big[p(\vu_{\leq m}) \mint \prod_{m'<m} t_{m'}(\vv_{m'})\dd \vu_{<m}\big]
     \big[\mint p(\vu_{\geq m}) \prod_{m'\geq m} t_{m'}(\vv_{m'}) \dd \vu_{>m}\big]/ p(\vu_m)\\
    &= q^f(\vu_{m})\,q^b(\vu_{m}) \,/\, p(\vu_m),\\ 
    q^s(\vv_m) 
    &= \mint q(\vu) \dd \vu_{\neq( m, m+1)}\\
    &= \mint p(\vu) \prod_{m'} t_{m'}(\vv_{m'}) \dd \vu_{\neq ( m, m+1)}\\
    &= \mint \big[ p(\vu_{\leq m}) \prod_{m'<m} t_{m'}(\vv_{m'})\big]
    t_m(\vv_m) p(\vu_{m+1}\mid \vu_m)
    \big[ p(\vu_{>m+1}\mid \vu_{m+1}) \prod_{m'\geq m+1} t_{m'}(\vv_{m'})\big] \dd \vu_{\neq ( m, m+1)}\\
    &=  \big[\mint p(\vu_{\leq m}) \prod_{m'<m} t_{m'}(\vv_{m'}) 
    \dd \vu_{<m} \big]
    t_m(\vv_m) p(\vu_{m+1}\mid \vu_m)/ p(\vu_{m+1})
    \big[\mint p(\vu_{\geq m+1}) \prod_{m'\geq m+1} t_{m'}(\vv_{m'})
    \dd \vu_{>m+1}\big] \\
    &=
q^f(\vu_{m})\,p(\vu_{m+1}\mid\vu_{m})\,t_m(\vv_{m}) \,/\, p(\vu_{m+1}) \,q^b(\vu_{m+1}).
\end{aligned}
\end{gather}
\paragraph{Kalman recursions} The above product of forward and backward filters is known as \emph{two-filter smoothing} \citep{sarkka2013bayesian}. An alternative way to implement this is via the more standard Kalman filter ($f$) and Rauch-Tung-Striebel (RTS) smoother ($s$). Letting $\vu_0=\vs(-\infty)$ and $\vu_{M+1}=\vs(\infty)$,
\begin{gather} \label{eq:kalman-recursions}
\begin{aligned}
    q^f(\vu_0) &= \NN(\vu_0 \mid \bm{0}, \vP_0), &&\text{\emph{initialise Kalman filter}} \\
    q^f(\vv_m) &\propto q^f(\vu_m) \, p(\vu_{m+1} \mid \vu_m) \, t_m(\vv_m), &&\text{\emph{compute joint, include site}} \\
    q^f(\vu_{m+1}) &= \mint q^f(\vv_m) \dd \vu_m, &&\text{\emph{marginalise (filtering dist.)}}\\
    q^s(\vu_M) &= \mint q^f(\vv_M) \dd \vu_{M+1}, &&\text{\emph{initialise RTS smoother}}\\
    q^p(\vu_{m+1}) &= \mint q^f(\vu_m) p(\vu_{m+1} \mid \vu_m) \dd \vu_{m}, &&\text{\emph{forward prediction}}\\
    q^s(\vu_m) &= q^f(\vu_m) \displaystyle\int \frac{ p(\vu_{m+1} \mid \vu_m) \, q^s(\vu_{m+1}) }{q^p(\vu_{m+1})} \dd \vu_{m+1}, \quad\quad\quad &&\text{\emph{smoothing dist.}}
\end{aligned}
\end{gather}
where $q^p(\cdot)$ is the forward filter prediction and $q^s(\cdot)$ is the desired smoothing distribution, \ie, the marginal posterior.

To derive the last line in \cref{eq:kalman-recursions} we let $\tilde{\vy}$ represent \emph{pseudo data} implied by the sites, $p(\tilde{\vy}_m \mid \vv_m) = t_m(\vv_m)$. With this notation the forward filter is given by $q^f(\vu_m) = p(\vu_m \mid \tilde{\vy}_{1:m}) \approx p(\vu_m \mid \vy_{1:n(m)})$, where $n(m)$ is the number of data points to the left of $z_m$, and the smoother by $q^s(\vu_m) = p(\vu_m \mid \tilde{\vy}_{1:M}) \approx p(\vu_m \mid \vy_{1:N})$, so we can write,
\begin{gather}
    \begin{aligned}
    q^s(\vu_m) &= p(\vu_m \mid \tilde{\vy}_{1:M}) \\
        &= \displaystyle\int p(\vu_{m}, \vu_{m+1} \mid \tilde{\vy}_{1:M}) \dd \vu_{m+1} && \\
        &= \displaystyle\int p(\vu_{m} \mid \vu_{m+1}, \tilde{\vy}_{1:M}) \, p(\vu_{m+1} \mid \tilde{\vy}_{1:M}) \dd \vu_{m+1} && \\
        &= \displaystyle\int p(\vu_{m} \mid \vu_{m+1}, \tilde{\vy}_{1:m}) \, q^s(\vu_{m+1}) \dd \vu_{m+1} && \\
        &= \displaystyle\int \frac{p(\vu_{m}, \vu_{m+1} \mid \tilde{\vy}_{1:m})}{p(\vu_{m+1} \mid \tilde{\vy}_{1:m})} \, q^s(\vu_{m+1}) \dd \vu_{m+1} && \\
        &= q^f(\vu_m) \displaystyle\int \frac{ p(\vu_{m+1} \mid \vu_m) \, q^s(\vu_{m+1}) }{q^p(\vu_{m+1})} \dd \vu_{m+1}.
    \end{aligned}
\end{gather}
\subsection{Normaliser}
\label{app:normaliser}

We are interested in the normalizer of $q(\vs)$.
A dense formulation can be obtained as follows:
\begin{align}
    \log \mint q(\vs(.)) \dd\vs
    &= \log \mint p(\vs(.)\mid \vu)\,p(\vu)\mprod_m t(\vv_m) \dd\vs\dd\vu \nonumber\\
    &= \log \mint p(\vu)\mprod_m t(\vv_m)\dd\vu \nonumber\\
    &= \log \mint \frac{e^{G(p(\vu))}\mprod_m z_m}{e^{G(q(\vu))}}q(\vu)\dd\vu \nonumber\\
    &= G(q(\vu)) - G(p(\vu)) + \msum_m \log\,z_m,
\end{align}
where we have defined the log-normaliser as the functional $G(\tNN(\vu; z, \vT_1, \vT_2)) =
\log \mint
\tNN(\vu; z, \vT_1, \vT_2)\dd\vu$.

A more efficient formulation dedicated to Markovian GPs is obtained using the filtering recursions of the previous section:
\begin{align}
    \mint q(\vu) \dd \vu &= \mint p(\vu)\,\mprod_m t_m(\vv_m) \dd \vu\nonumber\\
    &= \mint p(\vu_{m>1}\mid \vu_1) \mprod_{m>0} t_m(\vv_m)
    \underset{c^f_1\, q^f(\vu_1)}{\underbrace{\left[ 
    \mint p(\vu_1\mid \vu_0)\,t_0(\vv_0)\,p(\vv_0)\dd \vu_0
    \right]}} \dd \vu_{>0}\nonumber\\
    &= \mint p(\vu_{m>2}\mid \vu_2) 
    \mprod_{m>1} t_m(\vv_m)
    c^f_1 \underset{c^f_2 \,q^f(\vu_2)}{\underbrace{\left[\mint p(\vu_{2}\mid \vu_1)\,t_1(\vv_1)\,
    q^f(\vu_1) \dd \vu_1 \right]}} \dd \vu_{>1}\nonumber\\
    & = \dots = c^f_1 \dots c^f_{M-1} \mint  
    q^f(\vu_M) \dd \vu_{M} = c^f_1 \dots c^f_{M}.
\end{align}
The terms $c^f_m$ are the normalisers computed during the forward filtering recursions described in the previous section.
The normaliser can equivalently be computed using the backward filter:
\begin{align}
    \mint q(\vu) \dd \vu &= \mint p(\vu)\,\mprod_m t_m(\vv_m) \dd \vu\nonumber\\
    & = \mint  
    q^b(\vu_0) \dd \vu_{0} \, c^b_{0} \dots c^b_{M} = c^b_{0} \dots c^b_{M}.
\end{align}
Finally, the normaliser can also be computed using both the forward and backward filters, meeting at site indexed $m$:
\begin{align}
    \mint q(\vu) \dd \vu &= \mint p(\vu)\,\mprod_m t_m(\vv_m) \dd \vu\nonumber\\
    & =  c^f_1 \dots c^f_{m} \int  
    q^f(\vu_{m})\,p(\vu_{m+1}\mid\vu_{m})\, \,  \,\frac{q^b(\vu_{m+1})}
    {p(\vu_{m+1})} t_m(\vv_{m}) \dd \vv_{m} \, c^b_{m+1} \dots c^b_{M}.
\end{align}
This latter expression is useful when one needs to compute the normaliser of a site-based approximation after a single site update, as is the case in EP.

\section{Algorithms} \label{app:models}

\subsection{\ssvgp Algorithm} \label{app:ssvgp}

The approximate posterior process is parametrized as 
\begin{align}
    q(\vs(\cdot)) &= p(\vs(\cdot)\mid\vu)\,q(\vu) \nonumber \\
    &\propto p(\vs(\cdot)\mid\vu)  q(\vu_0)\mprod_{m=1}^M q_m(\vv_{m+1}\mid\vu_m). 
\end{align}
The variational lower bound to the marginal evidence is:
\begin{align} 
    \cL(q) = \EE_q \log p(\vy \mid f) - \text{KL}[q(\vu) \,\|\, p(\vu)].
\end{align}
The KL divergence is between two linear Gaussian state space models and thus decomposes as:
\begin{align} 
     \text{KL}[q(\vu) \,\|\, p(\vu)] = 
     \text{KL}[q(\vu_1) \,\|\, p(\vu_1)]
     + \msum_{m=1}^M \text{KL}\left[q(\vu_{m+1} \mid \vu_{m}) \,\|\, p(\vu_{m+1}\mid \vu_{m})\right].
\end{align}
Due to the locality of the conditional $f \mid \vu$, the variational expectation for a data point at $x$ such that $z_\om \leq x< z_\op$ is:
\begin{align}
q(f(x)) &= \mint p(f(x)\mid \vv_{m}=[\vu_{m}, \vu_{m+1}]) \, q(\vv_{m})\dd\vv_{m}\nonumber\\
&=   \mint  \NN(f(x) \mid \vW \vv, \nu) \, \NN(\vv_{m} \mid \vmu_{\vv_{m}}, \vSigma_{\vv_{m}})\dd\vv_{m}\nonumber \\
&=     \NN(f(x) \mid \vW \vmu_{\vv_{m}}, \vW \vSigma_{\vv_{m}} \vW^\top +\nu),
\end{align}
where $q(\vv_{m})$ is a pairwise posterior marginal over the consecutive inducing states $[\vu_{m}, \vu_{m+1}]$, which can be evaluated with linear time complexity in $M$, using classic Kalman smoothing algorithms (see \cref{app:filtering}).

\subsection{\sscvi Algorithm} \label{app:sscvi}

We follow the derivation of \cite{khan2017conjugate}. 
The approximate posterior process is parametrized using shared sites:
\begin{align}
    q(\vs(\cdot)) &= p(\vs(\cdot)\mid\vu)q(\vu) \nonumber \\
    &\propto p(\vs(\cdot)\mid\vu)  p(\vu)\prod_n t_m(\vv_m).
\end{align}
This is the same structure as for the \sspep algorithm, but here, since we approximate the posterior as a Gaussian, the normaliser of the sites are irrelevant.

The approximate posterior is optimized to get close to the true posterior in the sense of the KL divergence  $\text{KL}[q(\vs(\cdot)) \,\|\, p(\vs(\cdot) \mid \vy)]$, or equivalently by maximizing the variational objective:
\begin{align}
    \cL(q) = \EE_q \log p(\vy \mid f) - \text{KL}[q(\vu) \,\|\, p(\vu)].
\end{align}
The joint model is split into a conjugate and a non-conjugate part:
\begin{align}
    p(\vf, \vu, \vy) = \underbrace{p(\vu)}_{p_c(\vu)}\,\underbrace{p(\vf\mid\vu)\,p(\vy\mid\vf)}_{p_{nc}(\vf, \vu)}.
\end{align}
The conjugate part has \emph{sparse} minimal sufficient statistics $\phi(\vu)=[{(\vu_k, \vu_k \vu_k^\top)}^M_{k=1}, (\vu_{k+1} \vu_k^\top)_{k=1}^{M-1}]$, with the bilinear terms corresponding to the block-tridiagonal entries of matrix $\vu\vu^\top$ which we note $\text{btd}[\vu \vu^T]$.
We denote by $\vLambda$ the natural parameters of the prior $p(\vu)$ associated to sufficient statistics $\phi(\vu)$.

CVI approximates the non-conjugate part using Gaussian sites with the same sufficient statistics as the conjugate part: $\tilde{p}_{nc}(\vu) \approx p(\vf\mid\vu) t(\vu)$, where $t(\vu)=\prod_{m=1}^M t_m(\vv_m)$. Each site $t_m$ has natural parameter $\vlambda^{(m)}$ associated to local minimal sufficient statistics $\phi_m(\vu)=[\vv_m, \vv_m \vv_m^\top] \subset \phi(\vu)$. We denote by $\cP_m$ the linear operator projecting these minimal natural parameter into natural parameter with `full' sufficient statistics $\phi(\vu)$  and setting the rest of the natural parameters to $0$.
We denote by $\vlambda$ the projected natural parameters of the sites $\mprod_m t(\vv_m)$, \ie, $\vlambda= \msum_m \cP_m(\vlambda^{(m)})$. The natural parameters of the posterior over $\vu$ are thus $\vLambda + \vlambda$. 

One can show that a natural gradient step on the variational parameters $\vlambda^{(m)}$ boils down to \citep{khan2017conjugate}:
\begin{gather}\label{app:eq:cvi_update}
\begin{aligned}
    \vg^{(m)} &= \nabla_{\vmu^{(m)}} \EE_{q(\vf^{(m)})} \log p(\vy^{(m)}\mid\vf^{(m)})\\
    \vlambda^{(m)}_{k+1} &= (1-\rho)\, \vlambda^{(m)}_k + \rho\, \vg^{(m)}, 
\end{aligned}
\end{gather}
where $\vf^{m}$ and $\vy^{(m)}$ are here the subset of the data where the input $x$ falls in $[z_m, z_{m+1}]$, and $\vmu^{(m)}$ are the expectation parameters of the posterior $q(\vu^{(m)})$. %

These updates to the parameters can be written in terms of the derivatives of the variational expectations with respect to the mean and variance of the posterior marginal via the chain rule,
\begin{gather}
\begin{aligned} \label{eq:param-update-cvi}
    &\vg_{2}^{(m)} = \sum_{n \in \mathcal{M} } \vW_n^\top \diff{\mathcal{L}_n}{\Sigma_n} \vW_n, \\
    &\vg_{1}^{(m)} = \sum_{n \in \mathcal{M} } \vW_n^\top \diff{\mathcal{L}_n}{\mu_n} - 2 \vW_n^\top \diff{\mathcal{L}_n}{\Sigma_n} \vW_n \vmu_{m(n)},
\end{aligned}
\end{gather}
where $\mathcal{L}_n=\EE_{q(\vf^{(m)})} \log p(\vy^{(m)}\mid\vf^{(m)})$ and $\mu_n=\vW_n\vmu_{m(n)}$, $\Sigma_n=\vW_n \vSigma_{m(n)} \vW_n^\top + \nu_n$, and where $\mathcal{M}$ represents the indices to the data points whose inputs fall in $[z_m, z_{m+1}]$.

\subsubsection{\sscvi ELBO}
\label{app:sec:elbo}

Although the CVI method sidesteps direct computation of the ELBO for the variational parameter updates, it can still be used for hyperparameter learning. As in \cite{adam2020doubly}, the ELBO is given by:
\begin{align}\label{eq:elbo-f} 
\cL &= \EE_{q(\vs)} \log p(\vy \mid \vs) 
- \text{KL} \left[ q(\vu) \,\|\, p(\vu) \right] \end{align}
In \sscvi, we are interested in the normalized posterior, \ie, $q(\vu) = \mathcal{Z}^{-1} p(\vu)\mprod_m t_m(\vv_m)$, where $\mathcal{Z} = \mint p(\vu)\mprod_m t_m(\vv_m) \dd \vu$ is the normalizer (\ie, the marginal likelihood of the approximate conjugate model) and can be computed as shown in \cref{app:normaliser}.
The KL term in the  ELBO is:
\begin{align}\label{eq:elbo-kl}
\text{KL} \left[ q(\vu) \,\|\, p(\vu) \right] &=
\text{KL} \left[ \mathcal{Z}^{-1} p(\vu)\mprod_m t_m(\vv_m) \,\|\, p(\vu) \right] \nonumber \\
&= - \log \mathcal{Z}
+ \msum_m \EE_{q(\vv_m)}\log t_m(\vv_m) .
\end{align}
So the ELBO is:
\begin{align}\label{eq:elbo-f} 
\cL
&= \EE_{q(\vs)} \log p(\vy \mid \vs) 
+ \log \mathcal{Z}
- \sum_{m=1}^M \EE_{q(\vv_m)}\log t_m(\vv_m) .
\end{align}

\subsection{\sspep Algorithm} \label{app:sspep}

\setlength{\abovedisplayskip}{4pt}
\setlength{\belowdisplayskip}{4pt}

We follow the notation of \cite{bui2017unifying} in their derivation of the sparse PEP algorithm. There are two differences in our derivation: the latent process is an SDE, and the sites are inherently local due to the Markovian property of the model.
The starting point is a joint model of the data $\vy$ and the process prior $\vs(\cdot)$:
\begin{align}
    p(\vs(\cdot), \vy \mid \theta) = p(\vs(\cdot))\prod_{n=1}^{N}p(y_n\mid f_n, \theta).
\end{align}
In this setting, sparse EP consists of singling out a set of inducing inputs $\vZ= (z_1,\dots, z_M)\in \RR^M$ and using the associated inducing states $\vu= \vs(\vZ) \in \RR^{M\times d}$  to parametrize an approximation to this joint distribution of the form:
\begin{align}
    p(\vs(\cdot), \vy \mid \theta) \approx p(\vs(\cdot) \mid \vu) \, p(\vu) \prod_{n=1}^{N}t_n(\vu) = q(\vs(\cdot)),
\end{align}
where we denote $q(\vs(\cdot))$ to be the approximate \emph{joint}, which differs from the other algorithms we present.
The factors $t_n$ are called \emph{sites} and are parameterized as unnormalized Gaussian distributions in the natural parameterization: $t_n(\vu)= z_n \exp( \vu^\top \vT_{1,n} - \nicefrac{1}{2}\, \vu^\top \vT_{2,n}\vu) = \tNN(\vu; z_n, \vT_{1,n}, \vT_{2,n})$.

When there is one site per data point, the optimal form of the site is rank one: $t_n(\vu)= \tNN( \vW_n \vu; z_n, T_{1,n}, T_{2,n})$, where $\vW_n$ is the projection is the prior conditional mean $\EE_{p}[f_n\mid\vu] = \vW_n \vu$, and $z_n, T_{1,n}, T_{2,n}$ are scalars.

When working with Markovian GPs, the optimal site for data point $n$ can be shown to depend on the subset of inducing variables consisting of the two nearest inducing states $\vv_m(n) = [\vu_{m(n)}, \vu_{m(n)+1}]$, where $m(n)$ is such that $z_{m(n)} \leq x_n < z_{m(n)+1}$. So the final parameterization is $t_n(\vv_{m(n)})= \tNN( \vW_n \vv_{m(n)}; z_n, T_{1,n}, T_{2,n})$, where $\vW_n$ is the \emph{sparse} projection is the prior conditional mean ${\EE_{p}[f_n\mid\vu] =\EE_{p}[f_n\mid\vv_{m(n)}] = \vW_n \vv_{m(n)}}$.

Noting that all the $N_m$ data points whose input falls in $[z_m, z_{m+1}]$ have sites over $\vv_m$ makes those sites natural candidates to be \emph{locally tied} together: for each segment $[z_m, z_{m+1}]$, we replace each of the rank one sites $\{t_n(\vv_m); x_n \in [z_m, z_{m+1}]\}$ by a fraction of a full rank site $t_m(\vv_m)^{\nicefrac{1}{N_m}}$. Our approximation to the join thus becomes:
\begin{align}
     q(\vs(\cdot)) = p(\vs(\cdot) \mid \vu) \, p(\vu) \prod_{n=1}^{N}t_{m(n)}(\vv_{m(n)})^{\nicefrac{1}{N_{m(n)}}} = p(\vs(\cdot) \mid \vu) \, p(\vu) \prod_{m=0}^{M+1}t_m(\vv_m).
\end{align}

Given the above parametrisation, the \sspep algorithm involves three main steps: the cavity computation (`deletion'), moment matching (`projection'), and finally the update to the site parameters.

\subsubsection{Updates}
The three steps of the algorithm to update the sites are:
\begin{enumerate}
    \item \textbf{Deletion}: for a data point $n$, compute a cavity (which is an unnormalized Gaussian) by removing a fraction $k={\nicefrac{\alpha}{N_{m(n)}}}$ of a factor from the approximate joint $q(\vs(\cdot))$: 
    \begin{align}
        q^{\bl n}(\vs(\cdot)) = \frac{q(s(\cdot))}{t^{k}_{m(n)}(\vv_{m(n)})}.
    \end{align}
    This fraction $k$ can be understood as first picking the fraction of the shared site attributed to a data point ($1/N_{m(n)}$ where $N_{m(n)}$ is the number of sites tied together locally), then updating only a fraction $\alpha$ of this fraction.
    \item \textbf{Projection}: The new site is computed in the context of the other sites through the cavity, by minimizing the unnormalized KL divergence between the \emph{tilted} distribution $q^{\bl n}(\vs(\cdot)) \, p^{\alpha}(y_n\mid f_n)$ and the full approximate joint. Minimising the KL directly gives the new approximate joint $q^*(\vs(\cdot))$ as,
    \begin{align}
    q^{*}(\vs(\cdot)) \;
    \leftarrow \, \underset{q(\vs(\cdot)) \in \cQ}{\arg\min}\;
    \overline{\text{KL}}\left[q^{\bl n}(\vs(\cdot)) \, p^{\alpha}(y_n\mid f_n) \,{\Big\|}\, q(\vs(\cdot))\right].
    \end{align}
    Here, $\cQ$ is the set of acceptable distributions and corresponds to $\{q^{\bl n}(\vs(\cdot)) \, t^k(\vv_{m(n)}); \forall t\}$, in other words, the optimization only changes the site that has been removed to build the cavity.
    One can show that $q^{*}(\vv_{m(n)}) = \NN(\vv_{m(n)}\mid \vmu^\textrm{*}_{{m(n)}},\vSigma^\textrm{*}_{{m(n)}})$, where
    \begin{gather}
    \begin{aligned}
    \log Z_n &= \log \EE_{q^{\bl n}}[p^{\alpha}(y_n \mid f_n)],\\
    \vmu^\textrm{*}_{{m(n)}} &= \vmu_{m(n)} + \vW_{m(n)} \frac{\dd \log Z_n}{\dd \mu_n },\\
    \quad \vSigma^\textrm{*}_{{m(n)}} &= \vSigma_{m(n)} + \vW_{m(n)} \frac{\dd^2 \log Z_n}{\dd \mu_n^2 }\vW_{m(n)}^\top .
    \end{aligned}
    \end{gather}
    \item \textbf{Update}: Compute a new fraction of the approximate factor by dividing the new approximate joint by the cavity $t_{m(n),new}^{k}(\vv_{m(n)}) = q^{*}(\vv_{m(n)}) / q^{\bl n}(\vv_{m(n)})$ which is a rank one site.
    This fraction is then incorporated back to obtain the new site: $t^*_{m(n)}(\vv_{m(n)}) = t_{m(n),old}^{1-k}(\vv_{m(n)})t_{m(n),new}^{k}(\vv_{m(n)})$.\\
    The normaliser is then updated by matching the integral of the two terms in the KL divergence:
    \begin{gather}
    \begin{aligned}
        {\log \, \mint  q^{\bl n}(\vs(\cdot)) p^{\alpha}(y_n\mid f_n) \dd\vs(\cdot)}
        &={\log\, \mint q^{\bl n}(\vs(\cdot)) t^{k}_{m(n), new}(\vv_{m(n)})\dd\vs(\cdot)}\nonumber\\
        \log Z_n  &={\log\, \mint q^{\bl n}(\vu) t^{k}_{m(n), new}(\vv_{m(n)})\dd\vu}\\
        &=  G(q^*(\vu)) - G(q^{\bl n}(\vu)) + k \log z_{m(n), new},\\
        \log z_{m(n), new} &= \frac{1}{k}\left( \log Z_n -  G(q^*(\vu)) + G(q^{\bl n}(\vu))\right).
    \end{aligned}
    \end{gather}
    So the new site normaliser is $\log z^*_{m(n)} = (1-k)  \log\,z_{m(n), old}  + k \log\,z_{m(n), new}$.
    The normalizer can be computed efficiently using the recursions described in \cref{app:normaliser}
\end{enumerate}

\subsubsection{\sspep Energy}
\label{app:sec:pep_energy}

Following the approach of \citet{bui2017unifying}, the PEP energy is defined as the marginal likelihood of the approximate joint:
\begin{align}
    \log \cZ_{\text{PEP}} &= \log \mint q(\vs(.)) \dd\vs \nonumber\\
    &= \log \mint p(\vu)\,p(\vs(.)\mid \vu)\mprod_m t(\vv_m) \dd\vs\dd\vu \nonumber\\
    &= \log \mint p(\vu)\mprod_m t(\vv_m)\dd\vu \nonumber\\
    &= \log \int \frac{e^{G(p(\vu))}\mprod_m z_m}{e^{G(q(\vu))}}q(\vu)\dd\vu \nonumber\\
    &= G(q(\vu)) - G(p(\vu)) + \msum_m \log\,z_m,
\end{align}
This normalizer can be implemented efficiently as described in \cref{app:normaliser}.
It depends on the sites normalizer $z_m$ which themselves depend on the model hyper-parameters through the site update equations.
The energy function thus provides an objective to perform parameter optimization, as a proxy to the marginal likelihood $p(\vy)$.

We provide an alternative derivation of the same energy which is arguably easier to implement, and highlights the connection to the \sscvi ELBO. Recall that $t(\vv_{m})=\tNN(\vu; z_m, \vT_{1,m}, \vT_{2,m})=z_m\NN(\vu \mid \vT_{1,m}, \vT_{2,m})$, where $\vT_{1,m}$, $\vT_{2,m}$ are the natural parameters, then
\begin{align} 
\log \cZ_{\text{PEP}} &= \log \mint p(\vu)\mprod_m t(\vv_m)\dd\vu \nonumber\\
&= \log \int p(\vu) \mprod_m z_{m} \NN(\vv_{m} \mid \vT_{1,m}, \vT_{2,m}) \, \dd\vu \nonumber\\
&= \log \mprod_m z_m \int p(\vu) \mprod_m \NN(\vv_{m} \mid \vT_{1,m}, \vT_{2,m}) \, \dd\vu \nonumber\\
&= \log \mprod_m z_m + \log \int p(\vu) \mprod_m \NN(\vv_{m} \mid \vT_{1,m}, \vT_{2,m}) \, \dd\vu \nonumber\\
&= \msum_m \log z_m + \log \cZ ,
\end{align}
where $\log \cZ=\log \int p(\vu) \mprod_m \NN(\vv_{m} \mid \vT_{1,m}, \vT_{2,m}) \, \dd\vu$ is the normaliser of the approximate model, and can be computed in closed form via the Kalman filter as shown in \cref{app:normaliser}, or using the method in \cref{app:marglik}, replacing the true likelihood with $\NN(\vv_{m} \mid \vT_{1,m}, \vT_{2,m})$.

To compute $z_m$, the idea is to reuse the cavity computation, and to match the zero-th moment of the tilted distribution in the same way as we do for the first and second moments during inference. Let $\mathcal{Z}_{\text{lik},m}=\EE_{q_{\text{cav}}(\vv_{m})}[\mprod_{n \in \mathcal{M}} \EE_{p(f_n\mid \vv_{m})}[p^\alpha(y_n \mid f_n)]]=\mprod_{n\in \mathcal{M}} \EE_{q_{\text{cav}}(f_n)}[p^\alpha(y_n \mid f_n)]$, and $\mathcal{Z}_{\text{site},m}=\EE_{q_{\text{cav}}(\vv_{m})}[\NN^\alpha(\vv_m \mid \vT_{1,m}, \vT_{2,m})]$ be the cavity normalisers of the true likelihoods and the site approximations. We require the site constant factor, $z_m$, to be such that
\begin{gather}
\begin{aligned}
    z_m^\alpha \mathcal{Z}_{\text{site},m} &= \mathcal{Z}_{\text{lik},m} \\ 
    \implies z_m^\alpha &= \mathcal{Z}_{\text{lik},m} / \mathcal{Z}_{\text{site},m} \\
    \implies \log z_m &= \frac{1}{\alpha} (\log \mathcal{Z}_{\text{lik},m} - \log \mathcal{Z}_{\text{site},m}) ,
\end{aligned}
\end{gather}
so the full \sspep energy can be written,
\begin{align} 
\log \cZ_{\text{PEP}} &= \frac{1}{\alpha} \msum_m (\log \mathcal{Z}_{\text{lik},m} - \log \mathcal{Z}_{\text{site},m}) + \log \cZ \nonumber\\
&= \frac{1}{\alpha} {\msum_n\log \EE_{q_{\text{cav}}(f_n)}[p^\alpha(y_n \mid f_n)]} - \frac{1}{\alpha} \msum_m \log \EE_{q_{\text{cav}}(\vv_{m})}[\NN^\alpha(\vv_m \mid \vT_{1,m}, \vT_{2,m})] + \log \cZ.
\end{align}

\subsection{Approximate Marginal Likelihood via Approximate Filtering}\label{app:marglik}

The marginal likelihood can be expressed as,
\begin{align}
{p(\vy) = p(y_1)\mprod_{n=2}^N p(y_n \mid \vy_{1:n-1})}.
\end{align}
Further, each conditional term can be written (letting $\vs(x_n)=\vs_n$),
\begin{align}
p(y_n \mid \vy_{1:n-1})
= \mint p(y_n \mid f_n = \vH\vs_n) \, p(\vs_n \mid \vy_{1:n-1}) \dd\vs(x_n),
\end{align} 
where $p(\vs_n \mid \vy_{1:n-1})$ is the intractable forward filtering distribution:
\begin{align}
p(\vs_n \mid \vy_{1:n-1}) 
&= \mint p(\vs_n \mid \vs_{n-1}) \, p(\vs_{n-1} \mid \vy_{1:n-2}) \dd \vs_{n-1}. %
\end{align} 
Our approximation consists of running the approximate forward filter described in \cref{eq:kalman-recursions} to obtain $q^f(\vu_m)$ for $m=1,\dots,M$. We then approximate a single term $p(y_n \mid \vy_{1:n-1})$ as,
\begin{align}
    p(y_n \mid \vy_{1:n-1}) \approx \mint p(y_n \mid f_n) \, p(f_n \mid \vu_{m(n)}) \, q^f(\vu_{m(n)}) \, t^{k_n}(\vu_{m(n)}) \dd \vu_{m(n)},
\end{align}
where $t(\vu_{m(n)}) = \mint t_{m(n)}(\vv_{m(n)}) \dd \vu_{m(n)+1}$ is the contribution of the site in the forward direction and $k_n=N^{left}_{n}/N_{m(n)}$, with $N_{m(n)}$ being the number of data points whose inputs lie in $[z_{m(n)}, z_{m(n)+1}]$ and $N^{left}_{n}$ being the number of data points whose inputs lie in $[z_{m(n)}, x_n)$. Intuitively, this means the fraction of the site corresponding to the data points to the \emph{left} of $x_n$ are included. Here $f_n \mid \vu_{m(n)} \sim \NN(f_n \mid \vA_{m(n),n} \vmu_{m(n)}, \vA_{m(n),n} \vSigma_{m(n)} \vA_{m(n),n}^\top + \vQ_{m(n),n} )$.

\subsection{Posterior Linearisation (\sspl)}\label{app:slr}

In the general non-Gaussian likelihood case, when performing posterior linearisation we typically use the approximation $p(y_n \mid f_n) \approx \NN(\EE[y_n\mid f_n], \text{Cov}[y_n \mid f_n])$, allowing us to use the additive noise statistical linear regression (SLR) equations \citep{sarkka2013bayesian} in order to linearise the expected likelihood:
\begin{gather} \label{eq:gauss-filt-components-additive}
\begin{aligned}
\omega_n = \int & \EE[y_n\mid f_n] q(f_n) \, \textrm{d}f_n , \\
B_n = \int & \left[ (\EE[y_n\mid f_n] - \omega_n) (\EE[y_n\mid f_n] - \omega_n)^\top + \text{Cov}[y_n \mid f_n] \right] q(f_n)  \, \textrm{d}f_n , \\
C_n = \int & (f_n - \mu_n) (\EE[y_n\mid f_n] - \omega_n)^\top q(f_n) \, \textrm{d} f_n,
\end{aligned}
\end{gather}
where $\mu_n$ is the mean of the approximate marginal posterior $q(f_n)$.

As in \sscvi, the site updates for our extension to PL, \sspl, require only the posterior marginals, $q(f_n)$, whose moments are $\mu_n=\vW_n\vmu_{m(n)}$ and ${\Sigma_n=\vW_n\vSigma_{m(n)}^{-1}\vW_n^\top+\nu_n^2}$. 
The site update rule then proceeds as in \citet{wilkinson2020state}, but now including the projection back from $f_n$ to $\vv_{m(n)}$ through the conditional $f_n\mid\vv_{m(n)}$,
\begin{gather}
\begin{aligned} \label{eq:param-update-sl}
    &\vlambda_{2,n} = -\frac{1}{2}\vW_n^\top \Omega_n^\top \tilde{\Sigma}_n^{-1} \Omega_n \vW_n , \\
    &\vlambda_{1,n} = -2\vlambda_{2,n}\vmu_{m(n)} + \vW_n^\top \Omega_n^\top \tilde{\Sigma}_n^{-1} r_n .
\end{aligned}
\end{gather}
where we have introduced
\begin{gather}
\begin{aligned}
r_n&=y_n-\omega_n, \\
\tilde{\Sigma}_n&= B_n - C_n^\top \Sigma_n^{-1} C_n,\\
    \Omega_n &=\diff{\omega_n}{\mu_n} = \EE_{q(f_n)}\left[ \EE[y_n\mid f_n] \Sigma_n^{-1} (f_n - \mu_n)  \right].
\end{aligned}
\end{gather}

\paragraph{Extended Kalman Smoother (\sseks)} 
If the statistical linear regression equations are replaced by a first-order Taylor expansion, then PL reduces to the EKS. Hence we can also obtain a doubly sparse EKS (\sseks) algorithm by similarly substituting a Taylor expansion into the above. In practice, this amounts to setting $\tilde{\Sigma}_n=\text{Cov}[y_n\mid f_n]$ and $\Omega_n = \diff{\EE[y_n\mid f_n]}{f_n}|_{f_n=\mu_n}$. Whilst the EKS is not a common choice for modern day machine learning tasks, it does provide a useful trade off between efficiency, stability and performance. In particular, inference in \sseks avoids numerical integration, making it applicable in some scenarios where other methods are impractical.

\paragraph{PL Marginal Likelihood Approximation} When defining the PL marginal likelihood, \citet{garcia2019gaussian} assume a restrictive form for the sites, and discard a term in the marginal likelihood. However, the resulting approximation can be seen as a simplified form of the EP energy given in \cref{app:sec:pep_energy}. Therefore, to enable fair comparison, we use the EP energy for both \sspl and \sseks in all our experiments.

\section{Experimental Details}\label{app:experiment-details}

The following descriptions of our experimental tasks are adapted from \cite{wilkinson2020state}.

\paragraph{Motorcycle (heteroscedastic noise)}
The motorcycle crash data set \citep{Silverman:1985} contains 131 non-uniformly spaced measurements from an accelerometer placed on a motorcycle helmet during impact, over a period of 60~ms. It is a challenging benchmark \citep{Tolvanen:2014}, due to the heteroscedastic noise variance. We model both the process itself and the measurement noise scale with independent GP priors with Mat\'ern-$\nicefrac{3}{2}$ kernels: 
$y_n \mid f_n^{(1)},f_n^{(2)} \sim \NN(y_{n} \mid f^{(1)}(x_{n}),[\phi(f^{(2)}(x_n))]^2),$
with softplus link function $\phi(f) = \log(1+e^f)$ to ensure positive noise scale.

\paragraph{Coal (log-Gaussian Cox process)}
The coal mining disaster data set \citep{Vanhatalo+Riihimaki+Hartikainen+Jylanki+Tolvanen+Vehtari:2013} contains 191 explosions that killed ten or more men in Britain between 1851--1962. We use a log-Gaussian Cox process, \ie\ an inhomogeneous Poisson process (approximated with a Poisson likelihood for $N=333$ equal time interval bins). We use a Mat\'ern-$\nicefrac{5}{2}$ GP prior with likelihood 
$ p(\vy \mid \vf) \approx \prod_{n=1}^N \mathrm{Poisson}(y_n \mid \exp(f(\hat{x}_n)))$,
where $\hat{x}_n$ is the bin coordinate and $y_n$ the number of disasters in the bin. This model reaches posterior consistency in the limit of bin width going to zero \citep{Tokdar+Ghosh:2007}. For the linearisation-based inference methods (\sspl, \sseks) we utilise the fact that the first two moments are equal to the intensity, $\EE[y_n\mid f_n] = \text{Cov}[y_n \mid f_n] = \lambda(x_n) = \exp(f(x_n))$.

\paragraph{Airline (log-Gaussian Cox process)} 
The airline accidents data \citep{nickisch2018state} consists of 1210 dates of commercial airline accidents between 1919--2017. We use a log-Gaussian Cox process with bin width of one day, leading to $N = 35{,}959$ observations. The prior has multiple components, $\kappa(x,x') = \kappa(x,x')^{\nu=\nicefrac{5}{2}}_{\text{Mat.}}  + \kappa(x,x')_{\text{per.}}^{1\,\text{year}} \kappa(x,x')^{\nu=\nicefrac{1}{2}}_{\text{Mat.}} + \kappa(x,x')_{\text{per.}}^{1\,\text{week}} \kappa(x,x')^{\nu=\nicefrac{1}{2}}_{\text{Mat.}} $, capturing a long-term trend, time-of-year variation (with decay), and day-of-week variation (with decay). The state dimension is $d=59$.

\paragraph{Binary (1D classification)}
As a 1D classification task, we create a long binary time series, $N = 10{,}000$, using the generating function $y(x) = \text{sign} \{ \frac{12 \sin(4 \pi x)} {0.25 \pi x +1} + \sigma_x\}$, with $\sigma_x \sim \NN(0, 0.01^2)$. Our GP prior has a Mat\'ern-$\nicefrac{7}{2}$ kernel, $d=4$, and the sigmoid function $\psi(f) = (1 +e^{-f})^{-1}$ maps $\RR \mapsto [0,1]$ (logit classification).

\paragraph{Audio (product of GPs)} 
We apply a simplified version of the Gaussian Time-Frequency model from \citet{wilkinson2019end} to half a second of human speech, sampled at 44.1~kHz, $N=22{,}050$. The prior consists of 3 quasi-periodic ($\kappa_\textrm{exp}(x,x') \kappa_\textrm{cos}(x,x')$) `subband' GPs, and 3 smooth ($\kappa_\textrm{Mat-\nicefrac{5}{2}}(x,x')$) `amplitude' GPs. The likelihood consists of a sum of the product of these processes with additive noise and a softplus mapping $\phi(\cdot)$ for the positive amplitudes: $y_n \mid \vf_n \sim \NN(\sum_{i=1}^3 f_{i,n}^\textrm{sub.} \phi(f_{i,n}^\textrm{amp.}), \sigma^2_n)$. The nonlinear interaction of 6 GPs ($d=15$) in the likelihood makes this a challenging task.

In \cref{fig:vary_M} we analyse the effect of increasing the number of inducing inputs in the Audio task. We observe that the training marginal likelihood (NLML) and the test predictive density (NLPD) improve as $M$ increases, as expected for all methods. \sspep significantly outperforms the other methods, requiring fewer than 1000 inducing inputs to provide good results.

\begin{figure*}[h!]
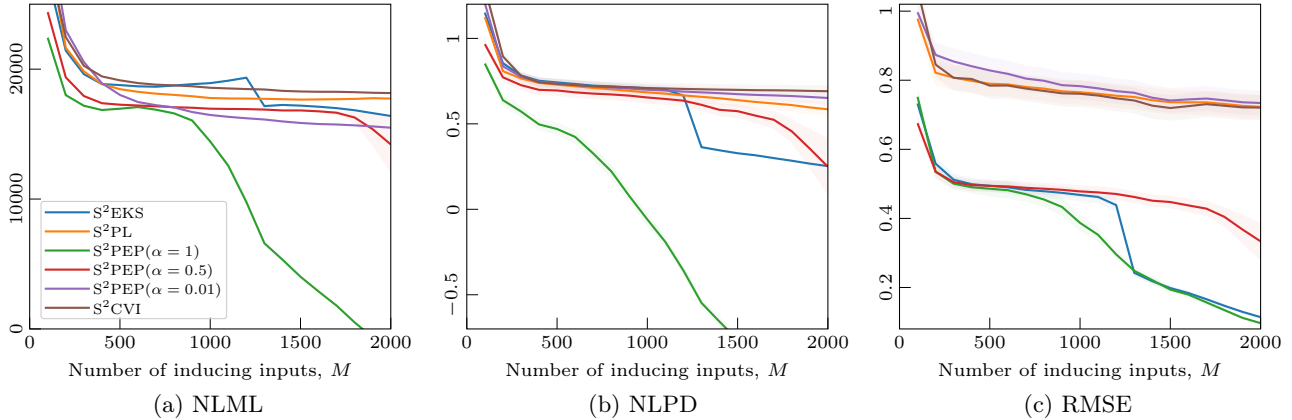

	\centering\scriptsize
	\tikzstyle{every picture}+=[remember picture]
	\pgfplotsset{scaled ticks=false,yticklabel style={rotate=90}, ylabel style={yshift=0pt},scale only axis,axis on top,clip=true,clip marker paths=true, xtick align=inside}
	\pgfplotsset{legend style={inner xsep=1pt, inner ysep=1pt, row sep=0pt},legend style={at={(0.98,0.98)},anchor=north east},legend style={rounded corners=1pt},legend style={fill=white, fill opacity=0.8, draw opacity=1, text opacity=1, draw=white!80!black},legend style={nodes={scale=0.8, transform shape}}}
	\setlength{\figurewidth}{.28\textwidth}
	\setlength{\figureheight}{0.9\figurewidth}
	\begin{subfigure}[b]{.32\textwidth}
	    \centering
	    \input{./graphs/audio_nlml.tex}
	    \vspace{-0.4cm}
	    \caption{NLML}
	\end{subfigure}
	\hfill
	\begin{subfigure}[b]{.32\textwidth}
	    \centering
	    \input{./graphs/audio_nlpd.tex}
	    \vspace{-0.4cm}
	    \caption{NLPD}
    \end{subfigure}
	\hfill
	\begin{subfigure}[b]{.32\textwidth}
	    \centering
	    \input{./graphs/audio_rmse.tex}
	    \vspace{-0.4cm}
	    \caption{RMSE}
	\end{subfigure}
	\vspace{-0.25cm}
  \caption{Analysis of the Audio task with varying number of inducing inputs. 
  \sspep($\alpha=1$) performs best in terms of test predictive density (NLPD) and RMSE, and requires many fewer inducing points. Whilst we expect \sspep($\alpha=0.01$) and \sscvi to give similar results, the numerical integration error when using 3-dimensional cubature causes the results to differ in practice.
  }
  \label{fig:audio_vary_M}
  \vspace{-0.25cm}
\end{figure*}

\paragraph{Banana (2D classification)}
The banana data set, $N=5300$, is a common 2D classification benchmark \citep{Hensman+Matthews+Ghahramani:2015}. We use the logit likelihood with a separable space-time kernel: $\kappa(r,x;r',x') =  \kappa(x,x')_{\text{Mat.}}^{\nu=\nicefrac{5}{2}} \kappa(r,r')^{\nu=\nicefrac{5}{2}}_{\text{Mat.}}$. The vertical dimension is treated as space, $r$, and the horizontal as the sequential (`temporal') dimension, $x$. We use $M=15$ inducing points in $r$, as well as $M=15$ inducing points in $x$. The state dimension is $d = 3M = 45$. For the SVGP baseline, we use $M=15^2=225$ inducing points placed on a 2D grid.

\paragraph{Electricity (large scale regression)}\label{app:electricity}We analyse the electricity consumption of one household \citep{Hebrail+Berard:2012,solin2018infinite} recorded every minute (in log~kW) over 1,442 days ($2{,}075{,}259$ total data points, with 25,979 missing observations). We assign the model a GP prior with a covariance function accounting for slow variation (Mat\'ern-$\nicefrac{3}{2}$) and daily periodicity with decay (quasi-periodic Mat\'ern-$\nicefrac{1}{2}$). We fit a GP to one 6 month's worth of data, which amounts to $N = 262{,}080$ points.

\end{document}